\documentclass[10pt,twocolumn,letterpaper]{article}

\usepackage[pagenumbers]{cvpr} 

\usepackage[accsupp]{axessibility} 
\usepackage[noend]{algpseudocode}
\usepackage{algorithm}
\definecolor{MScolor}{RGB}{238,105,131}
\definecolor{MIMcolor}{RGB}{91,155,213}
\definecolor{CLcolor}{RGB}{225,118,0}
\definecolor{RMcolor}{RGB}{0,32,96}
\definecolor{CRFR-Pcolor}{RGB}{255,0,0}
\usepackage{multirow}
\usepackage[normalem]{ulem}
\usepackage{colortbl}
\usepackage{amssymb}
\usepackage{hhline}
\usepackage{fontawesome5}

\definecolor{cvprblue}{rgb}{0.21,0.49,0.74}
\usepackage[pagebackref,breaklinks,colorlinks,allcolors=cvprblue]{hyperref}

\def\paperID{14775} 
\def\confName{CVPR}
\def\confYear{2025}

\title{FSFM: A Generalizable Face Security Foundation Model via Self-Supervised Facial Representation Learning}

\author{
  Gaojian Wang\textsuperscript{1,2},
  Feng Lin\textsuperscript{1,2}\thanks{Corresponding author}~,
  Tong Wu\textsuperscript{1,2},
  Zhenguang Liu\textsuperscript{1,2},
  Zhongjie Ba\textsuperscript{1,2},
  Kui Ren\textsuperscript{1,2}
  \\
  \textsuperscript{1}State Key Laboratory of Blockchain and Data Security, Zhejiang University\\
  \textsuperscript{2}Hangzhou High-Tech Zone (Binjiang) Institute of Blockchain and Data Security\\
  {\tt\small \{wolo, flin, cocotwu, liuzhenguang, zhongjieba, kuiren\}@zju.edu.cn}\\
  Project Page: \href{https://fsfm-3c.github.io/}{https://fsfm-3c.github.io}
}

\begin{document}
\maketitle
\begin{abstract}

This work asks: with abundant, unlabeled real faces, how to learn a robust and transferable facial representation that boosts various face security tasks with respect to generalization performance? We make the first attempt and propose a self-supervised pretraining framework to learn fundamental representations of real face images, \textbf{FSFM}, that leverages the synergy between masked image modeling (MIM) and instance discrimination (ID). We explore various facial masking strategies for MIM and present a simple yet powerful CRFR-P masking, which explicitly forces the model to capture meaningful intra-region \textbf{C}onsistency and challenging inter-region \textbf{C}oherency. Furthermore, we devise an ID network that naturally couples with MIM to establish underlying local-to-global \textbf{C}orrespondence through tailored self-distillation. These three learning objectives, namely \textbf{3C}, empower encoding both local features and global semantics of real faces. After pretraining, a vanilla ViT serves as a universal vision \textbf{F}oundation \textbf{M}odel for downstream \textbf{F}ace \textbf{S}ecurity tasks: cross-dataset deepfake detection, cross-domain face anti-spoofing, and unseen diffusion facial forgery detection. Extensive experiments on 10 public datasets demonstrate that our model transfers better than supervised pretraining, visual and facial self-supervised learning arts, and even outperforms task-specialized SOTA methods. 
\end{abstract}    
\section{Introduction}
\label{sec:intro}
The human face plays a crucial role in daily life and computer vision. In parallel, the facial security landscape suffers from diverse digital and physical manipulations, notably face forgery and presentation attacks. Face forgery edits digital pixels while keeping realistic visual quality. With significant advances in generative models~\cite{kingma2013auto, goodfellow2014generative, ho2020denoising}, the abuse of such technology, aka deepfakes, has led to severe trust crises. Presentation attacks utilize physical media, \eg, printed photos, image/video replays, or 3D masks, to spoof face recognition systems, risking real-life applications like face unlocking and payment. Thus, both academia and industry have been devoted to combating face forgery and presentation attacks with dedicated tasks: Deepfake Detection (DfD), Face Anti-Spoofing (FAS), and recently raised, Diffusion Facial Forgery detection (DiFF). 

Current DfD and FAS methods aim to improve generalization to unseen datasets within their respective tasks. DfD methods mainly focus on digital forgery patterns, such as spatiotemporal inconsistency~\cite{wang2023altfreezing, choi2024exploiting, bai2023aunet}, region artifacts~\cite{nguyen2024laa, hong2024contrastive, xia2024advancing}, and forgery augmentations~\cite{xia2024advancing, nguyen2024laa, bai2023aunet, shiohara2022detecting}. FAS methods~\cite{hu2024rethinking, zhou2024test, le2024gradient, zhou2023instance, sun2023rethinking} incorporate domain generalization techniques to discern physical spoof cues, \eg, paper textures and screen moiré patterns. Given these incompatible digital and physical features, most studies treat DfD and FAS as separate face security tasks, \ie, \textbf{task-specific} and \textbf{hard-to-transfer}. Besides, until now, most DfD and FAS methods still follow fully supervised learning with diverse backbones~\cite{chollet2017xception, tan2019efficientnet, dosovitskiy2020image, he2016deep} that are initialized from scratch or supervised ImageNet (natural images)~\cite{deng2009imagenet} pretraining. However, full supervision requires large-scale annotations or generative augmentations, which incur expensive costs and \textbf{limit scalability}. Moreover, initial weights that lack facial representations may \textbf{impede} \textbf{capability} and \textbf{generality} for face-related tasks~\cite{bulat2022pre, zheng2022general}.

In contrast to supervised learning, self-supervised learning (SSL) takes pretext tasks for pretraining on unlabeled data, where masked image modeling (MIM)~\cite{he2022masked, xie2022simmim, bao2021beit} and instance discrimination (ID, including contrastive learning~\cite{chen2020simple, he2020momentum, chen2021empirical} and distillation-based~\cite{grill2020bootstrap, chen2021exploring, caron2021emerging}) have proven superior performance in various vision tasks. As recent studies~\cite{park2023self, zhu2023understanding, ozbulak2023know} suggest that MIM and ID are complementary, SOTA SSL methods~\cite{tao2023siamese, assran2023self, huang2023contrastive, hernandez2025vic, chen2024context, zhao2024asymmetric} combine them to enhance visual representations for natural images. However, SSL progress for facial pretraining, especially security tasks, remains limited, raising \textbf{Q1: how can face security tasks benefit from self-supervised pretraining to learn scalable and generic representations?}

Although existing works introduce SSL to face security tasks, most focus on specific forgery or spoof patterns rather than transferable representations. In DfD, some methods~\cite{chen2022self, shiohara2022detecting, larue2023seeable, sun2025towards} synthesize forgeries from real faces to simulate artifacts like blending, vulnerable to unknown forgery or spoof types. Others~\cite{feng2023self, haliassos2022leveraging, zhao2022self} rely on audio-video datasets and multimodal information, limiting their scalability. A recent work~\cite{zhang2024learning} combines MIM and ID to learn temporal consistency~\cite{zhuang2022uia} from real face videos, but inter-frame representations fail for image-only forgeries and replay videos with consistent temporal features. Conversely, FAS methods employ SSL to mine spoof cues using domain positives~\cite{liu2023towards}, domain-invariant semantics~\cite{zheng2024mfae}, or domain alignment~\cite{liu2022source}, yet this FAS domain knowledge contributes less to detecting deepfake and diffusion forgery. 

Recent advances in facial pretraining~\cite{zheng2022general, cai2023marlin, bulat2022pre, liu2023pose, gao2024self} demonstrate that task-agnostic facial representations effectively transfer to several face analysis tasks, such as expression and attribute recognition. However, these methods focus primarily on salient facial features that forgery and spoof faces also exhibit well, \ie, overlook ``realness'' representations that go beyond normal facial analysis, thus hindering their extrapolation to face security tasks~\cite{zheng2022general}. Furthermore, these methods typically adopt intra-dataset evaluations for downstream tasks, whereas face security tasks call for cross-dataset generalization. These concerns raise \textbf{Q2: how to learn universal facial representations from real faces that transfer well to various face security tasks and improve downstream generalization?}

To bridge these gaps, instead of task-specific or unified~\cite{yuan2024unified, deb2023unified, yu2024benchmarking} supervised learning, we propose learning the intrinsic properties of unlabeled real face images through SSL. Specifically, we introduce a simple yet effective CRFR-P facial masking strategy into a masked autoencoder~\cite{he2022masked} that directs attention to pixel-level contexts and region-level correlations, yielding a meaningful and challenging MIM task. By Covering a Random Facial Region (\eg, eyes, nose) and Proportionally masking other regions, our CRFR-P promotes inter-region coherency and intra-region consistency in facial representations. For reliable face-level semantic alignment, we further integrate an ID network with MIM to establish local-to-global correspondence via elaborate self-distillation: the CRFR-P masked online view introduces spatial variance, the unmasked target view preserves complete semantics, and Siamese representation decoders form a disentangled space. Thus, our framework, FSFM, leverages the synergy between MIM and ID to empower both local and global perception, learns a universal representation of real faces, transfers well to various face security tasks, and achieves superior generalization performance. Our main contributions are:
\begin{itemize}
\item We propose a self-supervised pretraining framework to learn a fundamental representation of real faces, FSFM, which fully leverages the advantages of masked image modeling and instance discrimination for both local context perception and global semantic alignment.

\item We design a simple yet powerful CRFR-P facial masking strategy for MIM to enhance meaningful intra-region \textit{\textbf{C}onsistency} and capture challenging inter-region \textit{\textbf{C}oherency}, and we develop the ID network coupled with MIM to establish local-to-global \textit{\textbf{C}orrespondence} through elaborate self-distillation. Namely, \textit{\textbf{3C}}.

\item We use a vanilla ViT as the encoder to learn transferable and robust facial representations; after pretraining, it serves as a universal vision foundation model for various downstream face security tasks, while previous methods adopt diverse backbones that vary from specific tasks. 

\item We conduct extensive experiments on prevalent face security tasks involving 10 datasets: cross-dataset deepfake detection (DfD), cross-domain face anti-spoofing (FAS), and unseen diffusion facial forgery detection (DiFF), which demonstrate our model generalizes better than supervised practices, visual and facial self-supervised pretraining arts, and even outperforms task-specialized SOTA methods. Moreover, FSFM benefits from pretraining with more unlabeled real faces (a free lunch).

\end{itemize}

\section{Related work}

\begin{figure*}[t!]
\centering
\includegraphics[width=.95\textwidth]{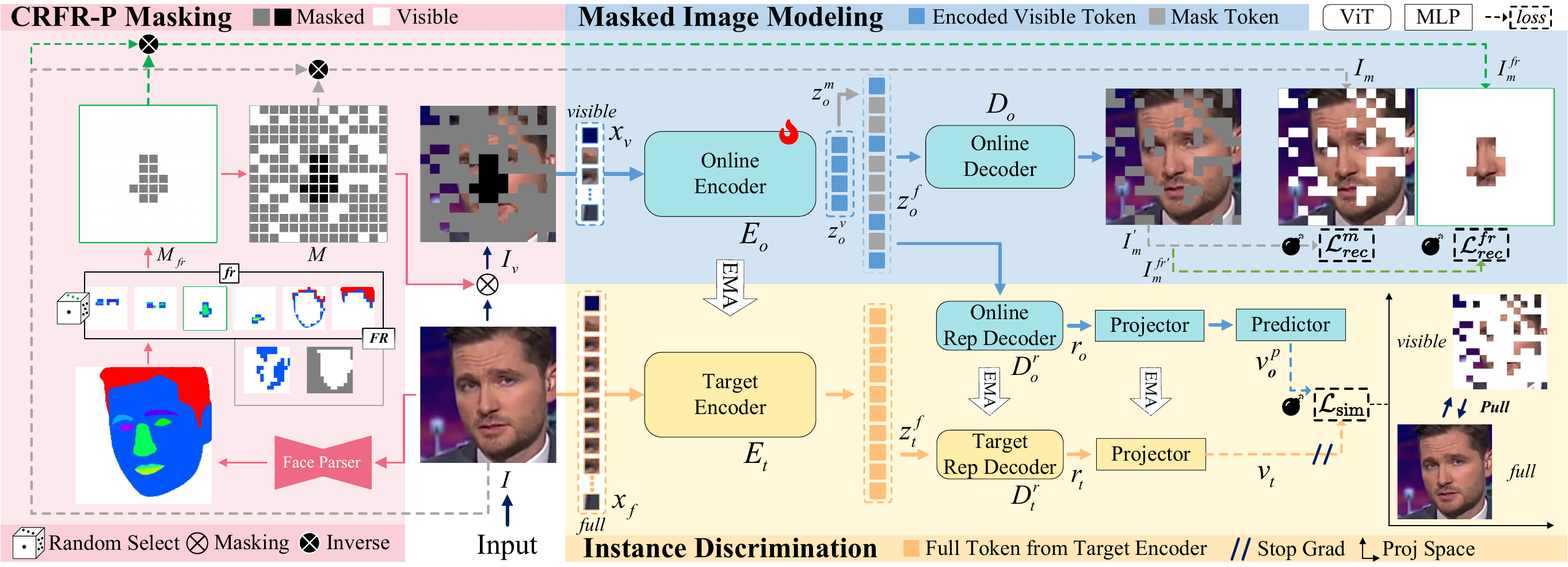}
\vspace{-5pt}
\caption{\textbf{Overview of FSFM} self-supervised pretraining framework for learning foundational representations of real faces (\textit{\textbf{3C}} \faBomb). Guided by the \textcolor{MScolor}{\textbf{CRFR-P masking}} strategy, the \textcolor{MIMcolor}{\textbf{masked image modeling (MIM)}} network captures \textit{intra-region \textbf{C}onsistency} with $\mathcal{L}_\mathit{rec}^\mathit{m}$ and enforces \textit{inter-region \textbf{C}oherency} via $\mathcal{L}_\mathit{rec}^\mathit{fr}$, while the \textcolor{CLcolor}{\textbf{instance discrimination (ID)}} network collaborates to promote \textit{local-to-global \textbf{C}orrespondence} through $\mathcal{L}_\mathit{sim}$. After pretraining, the online encoder $E_\mathit{o}$ (a vanilla ViT {\color{red} \faFire}) is applied to boost downstream face security tasks.}
\label{fig:framework}
\end{figure*}

\noindent\textbf{Visual representation learning} Masked image modeling (MIM) and instance discrimination (ID) are predominant pretext tasks for generative and discriminative self-supervised learning (SSL), respectively. MIM masks parts of an image and takes visible portions to restore the masked content, such as pixels~\cite{he2022masked, xie2022simmim}, visual tokens~\cite{bao2021beit}, or auxiliary features~\cite{wei2022masked}. With a high masking ratio, the tokenizer-free MAE~\cite{he2022masked} achieves efficient pretraining. Follow-up studies~\cite{li2022semmae, kakogeorgiou2022hide, shi2022adversarial, wang2023hard, nguyen2023r} highlight masking strategies and region-based learning. ID employs Siamese networks to pull positive pairs closer, mainly including contrastive learning (CL) and distillation (DIS) paradigms to avoid collapse. CL~\cite{chen2020simple, he2020momentum, chen2021empirical} simultaneously pushes negative pairs away. DIS~\cite{grill2020bootstrap, chen2021exploring, caron2021emerging} is negative-free and adopts asymmetric architecture to align representations between two branches. As recent studies~\cite{park2023self, zhu2023understanding, ozbulak2023know} suggest that MIM and ID complement each other, cutting-edge SSL methods increasingly combine MIM with ID via Siamese designs~\cite{assran2022masked, tao2023siamese, assran2023self, eymael2024efficient}, including CL~\cite{yi2022masked, huang2023contrastive, li2023mage, hernandez2025vic} and DIS~\cite{zhou2021ibot, chen2024context, bai2023masked, zhao2024asymmetric}, for natural vision representations.

\noindent\textbf{Facial representation learning} Several works~\cite{sun2024lafs, tourani2024pose, nguyen2023micron, wang2023ucol, li2020learning} explore facial SSL to reduce overfitting and improve performance, but they remain tailored for a single task, i.e., task-specific. VLP-based FaRL~\cite{zheng2022general} leverages web image-text CL with MIM to transfer across various facial tasks, but it requires image-text pairs and additional computation overhead, and the text captions tend to describe context rather than face details. Recent SSL efforts aim to learn general facial representations by MIM~\cite{cai2023marlin, wang2023toward}, CL~\cite{bulat2022pre,liu2023pose, wang2023toward}, DIS~\cite{gao2024self, wang2023toward} for various downstream tasks, such as expression~\cite{cai2023marlin, bulat2022pre, liu2023pose, gao2024self}, attribute~\cite{cai2023marlin, gao2024self}, and AU~\cite{bulat2022pre, liu2023pose} recognition, alignment~\cite{bulat2022pre, wang2023toward, gao2024self}, etc. Despite excelling in normal face analysis, these methods lack generality for face security tasks as they ignore facial ``realness'' representations. 
\section{FSFM framework}
\label{sec:framework}


We now describe the proposed FSFM pretraining framework in \cref{fig:framework}, which aims to learn intrinsic and transferable representations of real faces. The pretext tasks for self-supervised learning (SSL) include masked image modeling (MIM) and instance discrimination (ID). Driven by our CRFR-P masking strategy (\cref{subsec:CRFR-P}), the MIM network ($E_o \! \circ \! D_o$), a naive masked autoencoder (MAE)~\cite{he2022masked}, reconstructs the masked face to capture intra-region consistency and enforce inter-region coherency (\cref{subsec:MIM}). Meanwhile, the ID network employs the MIM encoder in the online branch ($E_o\circ D_o^r\circ \mathit{proj} \circ \mathit{pred}$) and establishes local-to-global correspondence, where the target branch ($E_t\circ D_t^r\circ \mathit{proj}$) guides global semantics (\cref{subsec:ID}). These three objectives, termed \textit{\textbf{3C}}, endow the online encoder with pixel-level context perceptiveness, region-level relation awareness, and instance-level face invariance. 

\subsection{CRFR-P masking}
\label{subsec:CRFR-P}
The mask sampling strategy plays a key role in MIM for both representation quality and downstream performance, and also provides the local view for our ID network. 

\begin{figure}[t!]
\centering
\includegraphics[width=1.\linewidth]{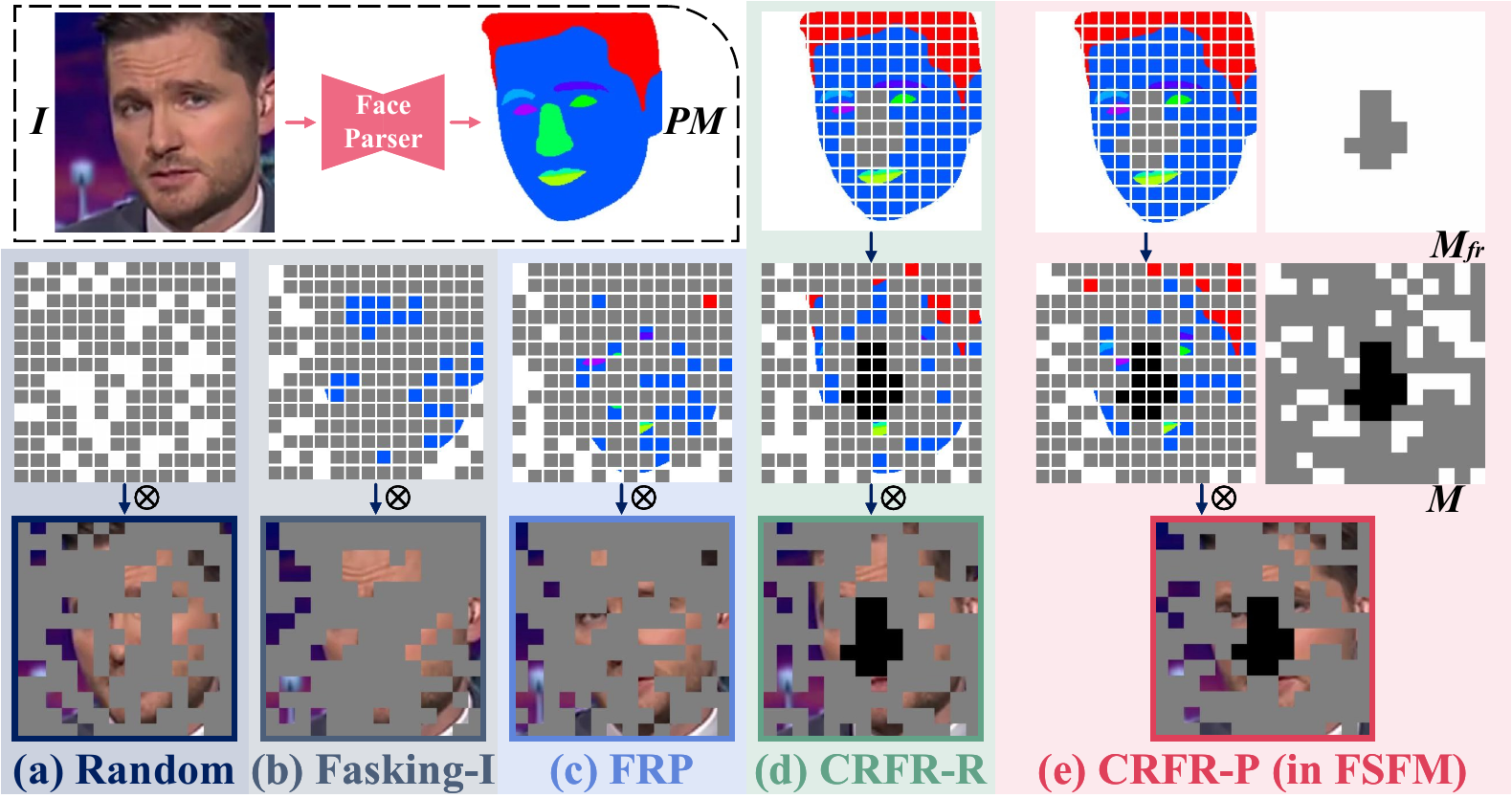}
\vspace{-17pt}
\caption{\label{fig:masking_strategies} Comparison of masking strategies for face images (75\% masking ratio). (a) Random masking. (b) Fasking-I, adapted from~\cite{cai2023marlin}, priority masking regions $\notin$\{bg, skin\}. (c) Our FRP: Proportional masking within each Facial Region $\in$\{$\mathit{FR}$\}. (d) Our CRFR-R: Covering a Random Facial Region $\in$\{$\mathit{fr}$\} and then Random masking other patches. (e) Our CRFR-P: Covering a Random Facial Region $\in$\{$\mathit{fr}$\} and then Proportional masking other regions $\in$\{$\mathit{FR}-\mathit{fr}$\}. All masks are binary (black solely highlights $\mathit{fr}$).}
\end{figure}

\noindent\textbf{Motivation} Random masking with a high mask ratio is widely used in natural~\cite{he2022masked, xie2022simmim} and facial~\cite{zheng2022general, wang2023toward} MIM, but it lacks domain-specific knowledge, limiting pretrained facial models. Given that human faces, our sole focus, comprise well-defined parts with distinct textures, we turn to explicitly segment the facial semantics for reasonable and efficient mask sampling rather than additional learning modules~\cite{li2022semmae, kakogeorgiou2022hide, shi2022adversarial, wang2023hard}. Similarly, MARLIN~\cite{cai2023marlin} uses an off-the-shelf face parser to divide facial parts and introduces the Fasking mask for face video MIM. We adapt Fasking to images as Fasking-I in \cref{fig:masking_strategies} (b): it divides facial parts into $\{$\textit{left-eye, right-eye, nose, mouth, hair, skin, background}$\}$, prioritizing masking non-skin and non-background regions. However, Fasking-I struggles to capture sufficient details solely from background and skin patches, rendering it unsuitable for face security tasks. Motivated by FACS~\cite{ekman1978facial} and psychological studies~\cite{russell1997psychology, haxby2000distributed}, we explore more effective masking strategies to learn local facial representations.

\noindent\textbf{Intuition} of intra-region consistency and inter-region coherency. As shown in \cref{fig:masking_strategies}, random masking and Fasking-I are prone to fully mask small but informative regions (\eg, eyes), hindering accurate learning of textures therein (\eg, pupils, iris). For intra-region consistency, we introduce FRP (Facial Region Proportional) masking that randomly masks an equal portion of patches in each region. FRP ensures all regions have visible patches, thus enabling attention to the same region when restoring masked patches. However, a potential shortcut, restoring a region's masked patches directly from its unmasked patches, may lead to a trivial reconstruction task and ignore cross-region correlations. In this regard, we formulate CRFR-R (Covering a Random Facial Region followed by Random) masking for inter-region coherency. As the fully masked region can only be derived from visible patches outside it, CRFR-R encourages learning its relationship \wrt other regions. Yet, the subsequent random masking may still obscure other small informative regions, undermining the desired consistency within them. Thus, FRP and CRFR-R, our preliminary masking strategies, have individual limitations but complement each other.

\noindent\textbf{CRFR-P Masking} Based on the above insights, we propose CRFR-P (Covering a Random Facial Region followed by Proportional) masking strategy, shown in \cref{fig:framework} and Alg. \ref{alg:CRFR_P}. CRFR-P first divides facial parts into predefined regions $\mathit{FR}$ using an off-the-shelf face parser. Next, it masks all patches within a randomly selected region $\mathit{fr}\notin$ \{\textit{skin, background}\} and obtains the facial region mask $M_\mathit{fr}$. Then, based on the number of masked patches and the overall masking ratio $r$, it randomly masks an equal portion of patches across the remaining $\{\mathit{FR}$$-$$\mathit{fr}\}$ regions to generate the image mask $M$. Finally, it returns both $M$ and $M_\mathit{fr}$. CRFR-P naturally resolves the core challenge: simultaneously promoting intra-region consistency and inter-region coherency. 

\noindent\textbf{Strength} Despite its simple design, CRFR-P masking poses a nontrivial and meaningful facial MIM task by effectively directing attention to critical facial regions with appropriate range and diversity. It not only avoids the trivial solution (shortcut) but also captures the intrinsic properties of real faces. In the supplementary material, we reveal key attention differences when applying different facial masking strategies to naive MAE pretraining.

\setlength{\textfloatsep}{15pt} 
\begin{algorithm}[t!]
\renewcommand{\thealgorithm}{1}
\caption{CRFR-P Masking Strategy}
\footnotesize  
{\bf Input:} Real face image $I$, Masking ratio $r$\\
{\bf Output:}  Image mask $M$, Facial region mask $M_\mathit{fr}$
\begin{algorithmic}[1]
\State $\mathit{PM} \gets \mathit{Face\_Parser}(I)$
\State $P_\mathit{pm} \in \mathbb{R}^N \gets \mathit{patchify}(\mathit{PM})$
\State $M, M_\mathit{fr} \gets [0] \in \mathbb{R}^N, [0] \in \mathbb{R}^N$
\State $\mathit{FR} \! \leftarrow \! \{$eyebrows \! $\supseteq \! [$right eyebrow, left eyebrow$]$, eyes \! $\supseteq \! [$right eye, left eye$]$, mouth \! $\supseteq \! [$upper lip, inner mouth, lower lip$]$, face boundary \! $\supseteq \! [$skin$\cap$background, skin$\cap$hair$]$, nose, hair, skin, background\}
\State Randomly select a $\mathit{fr} \in \{\mathit{FR} - \{\mathit{skin}, \mathit{background}\}\}$ 
\State $M_\mathit{fr}[P_\mathit{pm} \cap \mathit{fr}] \gets 1$ \Comment{\textit{\textbf{C}overing a \textbf{R}andom \textbf{F}acial \textbf{R}egion}}
\If{$\sum\! M_{\mathit{fr}} > N \cdot r$} \Comment{\textit{Extreme-case}}
    \State Randomly unmask $(\sum\! M_{\mathit{fr}} - N \cdot r)$ patches in $M_{\mathit{fr}}$
    \State $M \gets M_\mathit{fr}$
    \State \textbf{break}
\EndIf
\State \textbf{end if}
\State $M \gets M_\mathit{fr}$
\For{$pr \in \{\mathit{FR} - \{\mathit{fr}\}\}$} \Comment{\textit{\textbf{P}roportional masking in other regions}}
    \State $r=(N \cdot r-\sum\! M)~ / ~(N-\sum\! M)$
    \State $M[(P_\mathit{pm} \cap pr) \cdot r] \gets 1$
\EndFor
\State \textbf{end for}
\State \textbf{Return:} $M, M_\mathit{fr}$
\end{algorithmic}
\label{alg:CRFR_P}
\end{algorithm}

\subsection{MIM for facial region perception}
\label{subsec:MIM}
In a nutshell, the MIM network ($E_o \! \circ \! D_o$) in FSFM is an MAE~\cite{he2022masked} model guided by our CRFR-P masking strategy. Below, let $x_{f}=\left\{x_{i}\right\}_{i=1}^{N}$ denote the full set of $N$ non-overlapping patches split from the input face image $I$.

\noindent\textbf{Online Encoder} $E_{o}$ processes only on the visible patches $x_{v} \leftarrow M \odot x_{f}$ and maps $x_{v}$ into latent features $z_o^v$, where $\odot$ means the element-wise product for masking and $\leftarrow$ selects the visible ones. The online encoder first embeds $x_{v}$ by a linear projection as patch embeddings, and adds corresponding positional embeddings $p_{v}$. It then processes the fused embeddings through a series of transformer blocks to obtain: $z_o^v=E_o(x_v+p_v)$.

\noindent\textbf{Online Decoder} $D_{o}$ reconstructs the input image pixels. It first combines encoded visible tokens $z_o^v$ with mask tokens $z_o^m$, and adds relative positional embeddings to form the full tokens set $z_o^f$. The online decoder consists of another sequence of transformer blocks that receives $z_o^f$ as input, followed by a linear layer to restore the masked patches: $I_m^{\prime}=(1-M)\odot D_o(z_o^f)$.

\noindent\textbf{MIM Objective} Following~\cite{he2022masked}, we take normalized pixels as the reconstruction target and minimize the mean squared error (MSE) loss on masked patches between the prediction $I_{m}^{\prime}$ and the original one $I_m\leftarrow(1-M) \odot I$:
\begin{equation}
\setlength{\abovedisplayskip}{2pt}
\mathcal{L}_\mathit{rec}^m=\frac1{N_m}\sum\nolimits_{i=1}^{N_m}\left(I_m^{(i)}-I_m^{'(i)}\right)^2,
\setlength{\belowdisplayskip}{2pt}
\label{eq:loss_rec_m}\end{equation}
where $N_m$$=$$N$$\times$$r$$=$$\sum\! M$ is the number of masked patches. Our CRFR-P masking provides an additional mask $M_\mathit{fr}$, covering all patches in a randomly selected facial region $I_m^\mathit{fr}\leftarrow(1-M_\mathit{fr})\odot I$. To enhance inter-region coherency and prevent trivial solutions, we add another reconstruction loss only on the masked patches of the facial region $\mathit{fr}$:
\begin{equation}
\setlength{\abovedisplayskip}{2pt}
\mathcal{L}_\mathit{rec}^\mathit{fr}=\frac1{N_\mathit{fr}}\sum\nolimits_{j=1}^{N_\mathit{fr}}\left(I_m^\mathit{fr(j)}-I_m^{\mathit{fr}^{\prime}(j)}\right)^2,
\setlength{\belowdisplayskip}{2pt}
\label{eq:loss_rec_fr}
\end{equation}
where $N_\mathit{fr}$$=$$\sum\! M_\mathit{fr}$ is the number of patches in $\mathit{fr}$, and $I_m^{\mathit{fr}^{\prime}}\leftarrow(1-M_\mathit{fr})\odot I_m^{\prime}$. The overall MIM objective updates the MAE network ($E_o \! \circ \! D_o$) as a weighted sum:
\begin{equation}
\setlength{\abovedisplayskip}{2pt}
\mathcal{L}_\mathit{rec}=\mathcal{L}_\mathit{rec}^m+\lambda_\mathit{fr}\mathcal{L}_\mathit{rec}^\mathit{fr}.
\setlength{\belowdisplayskip}{2pt}
\label{eq:loss_mim}\end{equation}

\subsection{ID for local-to-global self-distillation}
\label{subsec:ID}

\begin{figure}[t!]
\centering
\includegraphics[width=\linewidth]{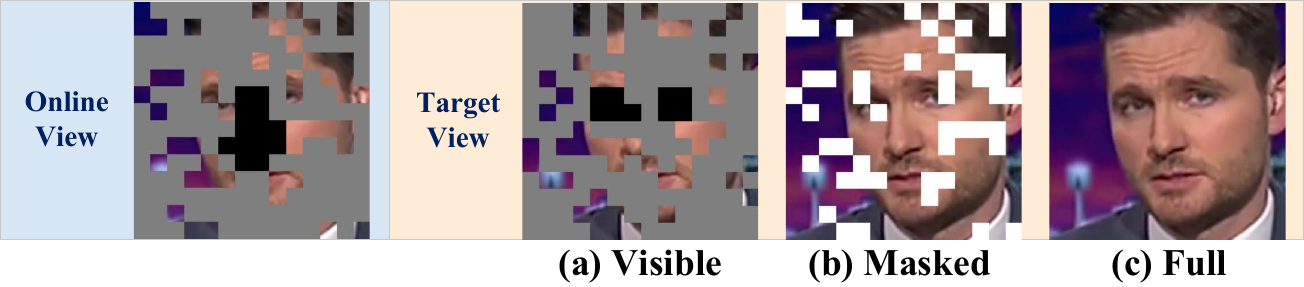}
\vspace{-17pt}
\caption{\label{fig:target_view} Comparison of different target views. (a) Visible patches from a different mask. (b) Masked patches from the same mask. (c) Full patches without masking.}
\end{figure}

In a nutshell, the ID network in FSFM employs symmetric designs between the online and target branches \wrt the encoder and representation decoder, as well as asymmetric designs \wrt the input view, projection head, negative-free loss, and model update rate. These designs focus on more precise and reliable semantic alignment for face security tasks, distinguishing our method from previous works that also integrate ID into MIM or degraded image input.

\noindent\textbf{Target Encoder} $E_{t}$ takes full patches $x_f$ as the target view to produce target latent features $z_t^f$ that guide the online encoder $E_{o}$ in learning holistic representations. Among the target view options in \cref{fig:target_view}, using (c) Full patches enables complete facial semantic embedding for local-to-global representation learning. Thus, the target encoder $E_{t}$ serves as a teacher that shares the same structure as the student $E_{o}$. Similarly, with positional embeddings $p_{f}$ of full patches, $E_{t}$ yields global embeddings: $z_t^f=E_t(x_f+p_f)$.

\noindent\textbf{Online and Target Rep Decoder} The online rep decoder $D_{o}^r$ maps the full tokens $z_o^f$ to online representations $r_o$. Unlike the online decoder $D_{o}$, which restores pixel values, $D_o^r$ reconstructs the masked token representations to align with the uncorrupted target, using significantly shallower transformer blocks followed by a linear layer that predicts features. The token features are output by a mean pooling as the online representations: $r_o=D_o^r(z_o^f)$.

In the target branch, the momentum encoder $E_{t}$ is updated by past iterations of the online encoder $E_{o}$, which also serves MIM. If $r_o$ is directly matched with target embeddings $z_t^f$, the model may struggle to fit high-level target features while restoring low-level pixels. Thus, a target rep decoder $D_{t}^r$ with the same structure as $D_{o}^r$ is added to represent the target features in the same disentangled space: $r_t=D_t^r(z_t^f)$.

\noindent\textbf{ID Objective} Following~\cite{grill2020bootstrap, chen2021exploring, chen2021empirical}, we use a projector and a predictor to transform $r_o$ into a lower-dimensional vector $v_o^p$, and apply only a projector to $r_t$ to obtain $v_t$. We then minimize the negative cosine similarity~\cite{chen2021exploring} between the $\ell_{2}$-normalized $v_o^p$ and $v_t$:
\begin{equation}
\setlength{\abovedisplayskip}{2pt}
\mathcal{L}_{sim}(v_o^p,\mathrm{sg}[v_t])=-\frac{v_o^p}{\|v_o^p\|_2}\cdot\frac{v_t}{\|v_t\|_2},
\setlength{\belowdisplayskip}{2pt}
\label{eq:loss_id}
\end{equation}
where $\mathrm{sg}[\cdot]$ is a stop-gradient, \ie, gradients are only calculated \wrt the online branch ($E_o\circ D_o^r\circ \mathit{proj} \circ \mathit{pred}$). The parameters $\theta_{t}$ of the target branch ($E_t\circ D_t^r\circ \mathit{proj}$) are updated via an exponential moving average (EMA)~\cite{grill2020bootstrap}: $\theta_t\leftarrow\tau\theta_t+(1-\tau)\theta_o$. Note that $\mathcal{L}_{sim}$ is asymmetric due to the different input views of the online and target branches, unlike the symmetrized designs~\cite{grill2020bootstrap, chen2021exploring, chen2021empirical, caron2021emerging}. 

\subsection{Overall pretraining objective}
\noindent\textbf{Overall Loss} FSFM learns foundational representations of real faces from both the MIM (\cref{eq:loss_mim}) and ID (\cref{eq:loss_id}) pretext tasks. Thus, the overall pretraining objective is a combined loss weighted by $\lambda_\mathit{cl}$:
\begin{equation}
\setlength{\abovedisplayskip}{2pt}
\mathcal{L}=\mathcal{L}_\mathit{rec}+\lambda_\mathit{cl}\mathcal{L}_\mathit{sim}\overset{\cref{eq:loss_mim}}{\operatorname*{=}}\mathcal{L}_\mathit{rec}^m+\lambda_\mathit{fr}\mathcal{L}_\mathit{rec}^\mathit{fr}+\lambda_\mathit{cl}\mathcal{L}_\mathit{sim}.
\setlength{\belowdisplayskip}{2pt}
\end{equation}
\section{Experiments}
\label{sec:exp}

In downstream face security tasks, including deepfake detection (DfD, \cref{subsec:exp_deeepfake}), face anti-spoofing (FAS, \cref{subsec:exp_fas}), and diffusion facial forgery detection (DiFF, \cref{subsec:exp_diff}), we demonstrate the effectiveness of FSFM by asking: \textbf{Q1:} Does our facial representation generalize better than common weight initialization practices? \textbf{Q2:} Is our method superior to other SSL methods for natural vision or normal facial analysis? \textbf{Q3:} Can our pretrained model outperform SOTA specialized methods with only simple finetuning? We further ablate FSFM (\cref{subsec:exp_ablation}) and present visualizations (\cref{subsec:exp_vis}) to ascertain our contributions. 
We include more pretraining, finetuning details, ablations, and visualizations in the supplementary material.

\begin{table*}[!t]
\centering
\resizebox{\linewidth}{!}{%
\begin{tabular}{lccccccc|lccccccc} \hline
\multirow{2}{*}{Method} & \multirow{2}{*}{\begin{tabular}[c]{@{}c@{}}Pretrain \\or Init\end{tabular}} & \multirow{2}{*}{\begin{tabular}[c]{@{}c@{}}Train\\Set\end{tabular}} & \multicolumn{4}{c}{Test Set \textbf{Video-level} AUC(\%)$\uparrow$} & \multirow{2}{*}{\begin{tabular}[c]{@{}c@{}}Avg.\\$\Delta$Ours\end{tabular}} & \multirow{2}{*}{Method} & \multirow{2}{*}{\begin{tabular}[c]{@{}c@{}}Pretrain \\or Init\end{tabular}} & \multirow{2}{*}{\begin{tabular}[c]{@{}c@{}}Train\\Set\end{tabular}} & \multicolumn{4}{c}{Test Set \textbf{Frame-level} AUC(\%)$\uparrow$} & \multirow{2}{*}{\begin{tabular}[c]{@{}c@{}}Avg.\\$\Delta$Ours\end{tabular}} \\ 
\cline{4-7}\cline{12-15}
\multicolumn{1}{c}{} &  &  & CDFV2 & DFDC & DFDCP & WDF &  & \multicolumn{1}{c}{} &  &  & CDFV2 & DFDC & DFDCP & WDF &  \\ 
\hline
\multicolumn{8}{l|}{\textbf{Base model}} & \multicolumn{8}{l}{\textbf{Base model}} \\
Xception~\cite{chollet2017xception} & Sup(IN) & FF++ & 76.39 & 70.62 & 72.24 & 76.11 & 14.0↑ & Xception~\cite{chollet2017xception} & Sup(IN) & FF++ & 69.52 & 68.20 & 68.94 & 68.83 & 15.1↑ \\
EfficientNet-B4~\cite{tan2019efficientnet} & Sup(IN) & FF++ & 79.81 & 71.85 & 66.95 & 76.42 & 14.1↑ & EfficientNet-B4~\cite{tan2019efficientnet} & Sup(IN) & FF++ & 73.37 & 69.47 & 64.37 & 71.95 & 14.2↑ \\
ViT-B~\cite{dosovitskiy2020image} & Scratch & FF++ & 64.08 & 66.73 & 72.62 & 60.36 & 21.9↑ & ViT-B~\cite{dosovitskiy2020image} & Scratch & FF++ & 61.14 & 64.27 & 69.00 & 60.68 & 20.2↑ \\
ViT-B~\cite{dosovitskiy2020image} & Sup(IN) & FF++ & 86.24 & 74.48 & 82.11 & 81.20 & 6.9↑ & ViT-B~\cite{dosovitskiy2020image} & Sup(IN) & FF++ & 77.43 & 71.09 & 74.07 & \uline{75.86} & 9.4↑ \\
MAE~\cite{he2022masked} ViT-B & SSL(IN) & FF++ & 79.51 & 75.93 & 87.10 & 80.96 & 7.0↑ & MAE~\cite{he2022masked} ViT-B & SSL(IN) & FF++ & 72.64 & 72.18 & 79.81 & 73.94 & 9.4↑ \\
DINO~\cite{caron2021emerging} ViT-B & SSL(IN) & FF++ & 80.47 & 76.90 & 84.64 & \uline{82.06} & 6.9↑ & DINO~\cite{caron2021emerging} ViT-B & SSL(IN) & FF++ & 73.88 & 72.78 & 77.31 & 75.08 & 9.2↑ \\
MCF~\cite{wang2023toward} ViT-B & SSL(LFc) & FF++ & 80.25 & 73.61 & 82.55 & 79.79~ & 8.8↑ & MCF~\cite{wang2023toward} ViT-B & SSL(LFc) & FF++ & 73.16 & 69.63 & 75.78 & 74.10 & 10.8↑  \\
\rowcolor[rgb]{0.8,0.8,0.8} \textbf{FSFM ViT-B~\textbf{(Ours)}} & SSL(VF2) & FF++ & 91.44 & \textbf{83.47} & \textbf{89.71} & \textbf{86.96} & - & \textbf{FSFM ViT-B (Ours)} & SSL(VF2) & FF++ & \textbf{85.05} & \textbf{80.20} & \textbf{85.50} & \textbf{85.26} & - \\ 
\hline
\multicolumn{8}{l|}{\textbf{SOTA specialized method (Venue)}} & \multicolumn{8}{l}{\textbf{SOTA specialized method (Venue)}} \\
SBIs~\cite{shiohara2022detecting} (CVPR'22)‡ & Init(IN) & SD & 93.18 & 72.42 & 86.15 &  & 4.3↑ & OST~\cite{chen2022ost} (NIPS'22)† & Init(IN) & FF++ & 74.80 &  & \uline{83.30} &  & 6.2↑ \\
RealForensics~\cite{haliassos2022leveraging} (CVPR'22) & SSL(LRW) & FF++ & 86.90 & 75.90 &  &  & 6.1↑ & FInfer~\cite{hu2022finfer} (AAAI'22)† & Init(IN) & FF++ & 70.60 &  & 70.39 & 69.46 & 15.1↑ \\
HCIL~\cite{gu2022hierarchical} (ECCV'22) & Init(IN) & FF++ & 79.00 &  & 69.21 &  & 16.5↑ & PEL~\cite{gu2022exploiting} (AAAI'22)‡ & Init(IN) & FF++ & 69.18 & 63.31 &  & 67.39 & 16.9↑ \\
 SeeABLE~\cite{larue2023seeable} (ICCV'23)‡ & SSL(SD) & SD & 87.30 & 75.90 & 86.30 &  & 5.0↑ & SLADD~\cite{chen2022self} (CVPR'22)† & Init(IN) & SD & 79.70 &  & 81.80 &  & 4.5↑ \\
TALL~\cite{xu2023tall} (ICCV'23)\textbf{*} & Init(IN) & SD & 90.79 & 76.78 &  &  & 3.7↑ & RECCE~\cite{cao2022end} (CVPR'22)† & Init(IN) & FF++ & 68.71 & 69.06 &  & 64.31 & 16.1↑ \\
AUNet~\cite{bai2023aunet} (CVPR'23)† & Init(IN) & SD & 92.77 & 73.82 & 86.16 &  & 4.0↑ & UIA-ViT~\cite{zhuang2022uia} (ECCV'22)\textbf{*} & SSL(FF) & FF++ & 82.41 &  & 75.80 &  & 6.2↑ \\
SLF~\cite{choi2024exploiting} (CVPR'24) & FTCN & FF++ & 89.00 &  &  &  & 2.4↑ &  UAL~\cite{wu2023generalizing} (MM'23)\textbf{*} & Init(IN) & FF++ & 82.84 &  &  & 70.13 & 8.7↑ \\
MLR~\cite{hong2024contrastive} (CVPR'24)\textbf{*} &  & FF++ & 91.56 & 75.17 &  & 73.41 & 7.2↑ & NoiseDF~\cite{wang2023noise} (AAAI'23) & RIDNet & FF++ & 75.89 &  & 63.89 &  & 15.3↑ \\
LAA-Net/BI~\cite{nguyen2024laa} (CVPR'24)‡ & Init(IN) & SD & 86.28 &  & 69.69 & 57.13 & 18.3↑ & GS~\cite{guo2023controllable} (ICCV'23)‡ &  & FF++ & \uline{84.97} &  & 81.65 &  & 1.8↑ \\
LAA-Net/SBI~\cite{nguyen2024laa} (CVPR'24)‡ & Init(IN) & SD & \textbf{95.40} &  & 86.94 & 80.03 & 1.9↑ & UCF~\cite{yan2023ucf} (ICCV'23)† & Init(IN) & FF++ & 82.40 &  & 80.50 &  & 3.7↑  \\
LSDA~\cite{yan2024transcending} (CVPR'24)‡ & Init(IN) & FF++ & 91.10 & \uline{77.00} &  &  & 3.4↑ & SFDG~\cite{wang2023dynamic} (CVPR'23)‡ & Init(IN) & FF++ & 75.83 & \uline{73.64} &  & 69.27 & 10.6↑ \\
FPG~\cite{xia2024advancing} (MM'24)‡ & Init(IN) & SD & \uline{94.49} & 74.75 & \uline{87.24} &  & 2.7↑ & IID~\cite{huang2023implicit} (CVPR'23) & CF(WF) & FF++ & 83.80 &  & 81.23 &  & 2.8↑ \\
NACO~\cite{zhang2024learning} (ECCV'24)\textbf{*} & SSL(VC2) & FF++ & 89.50 & 76.70 &  &  & 4.4↑ & LSDA~\cite{yan2024transcending} (CVPR'24)‡ & Init(IN) & FF++ & 83.00 & 73.60 & 81.50 &  & 4.2↑ \\
\hline
\multicolumn{16}{l}{\textit{\textbf{Abbreviation:~~}}Sup-Supervised~ SSL-Self-Supervied~ Init-weight initialization~ ~ \textit{\textbf{Dataset:~~}}IN-ImageNet~ LRW-LipReading in the Wild/facial~ VC2-VoxCeleb2/facial~ CF(WF)-CosFace(WebFace)/facial} \\
\multicolumn{16}{l}{SD-Synthetic (or Self-made) Data~ ~~\textbf{\textit{Backbone(or modified):~~~~}}† Xception~ ~~‡ EfficientNet-B4~ ~ \textbf{*} ViT or transformer-based}
\end{tabular}
}%
\vspace{-5pt}
\caption{Cross-dataset evaluation on deepfake detection (DfD). Left: video-level, Right: frame-level. All base models are finetuned on FF++ (c23) and tested on unseen datasets. For a fair comparison, the results of SOTA specialized methods are cited from their original papers. Avg.$\Delta$Ours denotes the average AUC difference between our FSFM and other methods on their test sets. \textbf{Best results}, \underline{second-best}.}
\label{tab:df_compare}
\end{table*}

\subsection{Pretraining setups and baselines}
\noindent\textbf{Pretraining Data and Preprocessing} We use the VGGFace2~\cite{cao2018vggface2} dataset (VF2, $\sim$3.3M images) for pretraining. We utilize the DLIB~\cite{king2009dlib} for face detection and cropping with a 30\% addition margin, and the FACER~\cite{zheng2022general} toolkit for face parsing instead of alignment. Cropped face images are resized to 224×224, and parsing maps are saved as binary stream files for efficient CRFR-P masking. 

\noindent\textbf{Model Architecture} Our MIM network is a vanilla MAE~\cite{he2022masked}  with ViT-B/16 as the default encoder $E_o$. In the ID network: rep decoders, $D_{o}^r$ and $D_{t}^r$ are 2-layer ViT blocks with the same width as the encoder; the projector and predictor are 2-layer MLPs following BYOL~\cite{grill2020bootstrap}. After pretraining, we only use $E_o$ as the backbone and add a task-specific head for downstream finetuning.

\noindent\textbf{Implementation} We set the mask ratio $r$ to 0.75. We use no data augmentation (not even crop or flip used in ~\cite{he2022masked}) during pretraining. We empirically set the loss weights $\lambda_\mathit{fr}$, $\lambda_\mathit{cl}$ to 0.007, 0.1, respectively. The EMA momentum coefficient follows~\cite{grill2020bootstrap}. We pretrain our model from scratch for 400 epochs. Other setups follow MAE~\cite{he2022masked} defaults.

\noindent\textbf{Pretrained Base Models} To answer \textbf{Q1} and \textbf{Q2}, we include the following pretrained models based on availability (official weights), fairness (vision-only ViT-B/16 backbone), and relevance: 1) ViT-B Scratch~\cite{dosovitskiy2020image}, to discern pretraining effects versus the backbone itself; 2) ViT-B Sup(IN)~\cite{dosovitskiy2020image}, supervised ImageNet pretraining, the most common weight initialization for face security tasks; 3) MAE SSL(IN)~\cite{he2022masked}, our MIM network, the baseline across all experiments, including ablations; 4) DINO SSL(IN)~\cite{caron2021emerging}, a self-distillation method for ID, which also encourages local-to-global correspondence; 5) MCF SSL(LFc)~\cite{wang2023toward}, a SOTA facial representation model for face analysis tasks, which also combines MIM and ID and is pretrained on the large-scale LAION-FACE-cropped dataset ($\sim$20M facial images). For downstream tasks, we compare our FSFM with these models by initializing the backbone with their pretrained weights and keeping other settings identical. We employ end-to-end finetuning for all models, facilitating a fair comparison with SOTA specialized methods to answer \textbf{Q3}.

\subsection{Deepfake detection}
\label{subsec:exp_deeepfake}

\begin{table*}[!t]
\centering
\resizebox{\linewidth}{!}{
\begin{tabular}{p{5.2cm}p{1.8cm}<{\centering}p{0.4cm}<{\centering}p{0.4cm}<{\centering}p{0.4cm}<{\centering}p{0.4cm}<{\centering}p{0.4cm}<{\centering}p{1.3cm}<{\centering}p{1.3cm}<{\centering}p{1.3cm}<{\centering}p{1.3cm}<{\centering}p{1.3cm}<{\centering}p{1.3cm}<{\centering}p{1.3cm}<{\centering}p{1.3cm}<{\centering}p{1.3cm}<{\centering}p{1.3cm}<{\centering}}
\hline
\multirow{2}{*}{Method} & \multirow{2}{*}{\begin{tabular}[c]{@{}c@{}}Pretrain\\or Init\end{tabular}} & \multicolumn{5}{c}{DG FAS Technique} & \multicolumn{2}{c}{OCI→M} & \multicolumn{2}{c}{OMI→C} & \multicolumn{2}{c}{~OCM→I} & \multicolumn{2}{c}{ICM→O} & \multicolumn{2}{c}{Avg.} \\ 
\cline{3-17}
 &  & DM & AL & CL & ML & PL & HTER↓ & AUC↑ & HTER↓ & AUC↑ & HTER↓ & AUC↑ & HTER↓ & AUC↑ & HTER↓ & AUC↑ \\ 
\hline
\multicolumn{17}{l}{\textbf{Base model}} \\
ViT-B~\cite{dosovitskiy2020image} & Scratch &  &  &  &  &  & 15.37 & 90.73 & 35.37 & 68.23 & 14.75 & 94.18 & 31.65 & 71.55 & 24.28 & 81.17~ \\
ViT-B~\cite{dosovitskiy2020image}(@ECCV'22~\cite{huang2022adaptive}) & Sup(IN) &  &  &  &  &  & \textbf{3.52} & 98.74 & \textbf{2.42} & \textbf{99.52} & 8.45 & 96.91 & 11.86 & 94.62 & 6.56 & \uline{97.44~} \\
MAE~\cite{he2022masked} ViT-B & SSL(IN) &  &  &  &  &  & 10.32 & 94.87 & 15.91 & 89.96 & 15.54 & 91.13 & 16.51 & 90.29 & 14.57 & 91.56~ \\
DINO~\cite{caron2021emerging} ViT-B & SSL(IN) &  &  &  &  &  & 6.73 & 97.15 & 13.44 & 93.90 & 14.27 & 93.56 & 15.55 & 90.99 & 12.50 & 93.90~ \\
MCF~\cite{wang2023toward} ViT-B & SSL(LFc) &  &  &  &  &  & 4.00 & \uline{98.84} & 8.46 & 96.90 & 8.02 & 97.39 & 10.70 & 95.64 & 7.80 & 97.19~ \\
\rowcolor[rgb]{0.8,0.8,0.8} \textbf{FSFM ViT-B (Ours)} & SSL(VF2) &  &  &  &  &  & \uline{3.78} & \textbf{99.15} & \uline{3.16} & \uline{99.41} & \uline{4.63} & \textbf{99.03} & \textbf{7.68} & 97.11 & \textbf{4.81} & \textbf{98.68~} \\ 
\hline
\multicolumn{17}{l}{\textbf{SOTA specialized method (Venue)}} \\
MADDG~\cite{shao2019multi} (CVPR'19)† & Scratch & \checkmark & \checkmark & \checkmark &  &  & 17.69 & 88.06 & 24.50 & 84.51 & 22.19 & 84.99 & 27.98 & 80.02 & 23.09 & 84.40~ \\
SSDG-R~\cite{jia2020single} (CVPR'20)\textsuperscript{\#} & Init(IN) &  & \checkmark & \checkmark &  &  & 7.38 & 97.17 & 10.44 & 95.94 & 11.71 & 96.59 & 15.61 & 91.54 & 11.29 & 95.31~ \\
AMEL~\cite{zhou2022adaptive} (MM'22)† & Scratch & \checkmark &  &  & \checkmark &  & 10.23 & 96.62 & 11.88 & 94.39 & 18.60 & 88.79 & 11.31 & 93.96 & 13.01 & 93.44~ \\
FGHV~\cite{liu2022feature} (AAAI'22)‡ & Depth Net &  & \checkmark & \checkmark &  &  & 9.17 & 96.92 & 12.47 & 93.47 & 16.29 & 90.11 & 13.58 & 93.55 & 12.88 & 93.51~ \\
PatchNet~\cite{wang2022patchnet} (CVPR'22)\textsuperscript{\#} & Init(IN) &  &  & \checkmark &  &  & 7.10 & 98.46 & 11.33 & 94.58 & 13.40 & 95.67 & 11.82 & 95.07 & 10.91 & 95.95~ \\
SSAN-R~\cite{wang2022domain} (CVPR'22)\textsuperscript{\#} & Init(IN) & \checkmark & \checkmark & \checkmark &  &  & 6.67 & 98.75 & 10.00 & 96.67 & 8.88 & 96.79 & 13.72 & 93.63 & 9.82 & 96.46~ \\
UDG-FAS~\cite{liu2023towards} (ICCV'23)\textsuperscript{\#} & SSL(LOO) &  &  & \checkmark &  &  & 7.14 & 97.31 & 11.44 & 95.59 & 6.28 & 98.61 & 12.18 & 94.36 & 9.26 & 96.47~ \\
UDG-FAS~\cite{liu2023towards} (ICCV'23)\textsuperscript{\#} & Init(IN) &  &  & \checkmark &  &  & 5.95 & 98.47 & 9.82 & 96.76 & 5.86 & 98.62 & 10.97 & 95.36 & 8.15 & 97.30~ \\
IADG~\cite{zhou2023instance} (CVPR'23)† & Scratch & \checkmark &  & \checkmark &  &  & 5.41 & 98.19 & 8.70 & 96.44 & 10.62 & 94.50 & 8.86 & \uline{97.14} & 8.40 & 96.57~ \\
SAFAS~\cite{sun2023rethinking} (CVPR'23)\textsuperscript{\#} & Init(IN) &  &  & \checkmark &  &  & 5.95 & 96.55 & 8.78 & 95.37 & 6.58 & 97.54 & 10.00 & 96.23 & 7.83 & 96.42~ \\
GAC-FAS~\cite{le2024gradient} (CVPR'24)\textsuperscript{\#} & Init(IN) &  &  &  &  &  & 5.00 & 97.56 & 8.20 & 95.16 & \textbf{4.29} & \uline{98.87} & \uline{8.60} & \textbf{97.16} & \uline{6.52} & 97.19~ \\
HPDR~\cite{hu2024rethinking} (CVPR'24)‡ & Depth Net &  &  &  &  & \checkmark & 4.58 & 96.02 & 11.30 & 94.42 & 11.26 & 92.49 & 9.93 & 95.26 & 9.27 & 94.55~ \\
TTDG~\cite{zhou2024test} (CVPR'24)† & Scratch & \checkmark &  & \checkmark &  &  & 7.91 & 96.83 & 8.14 & 96.49 & 6.50 & 97.98 & 10.00 & 95.70 & 8.14 & 96.75~ \\ 
\hline
\multicolumn{16}{l}{\textit{\textbf{DG (Domain Generalization) FAS Technique:~~}}DM-Depth Maps~ AL-Adversarial Learning~ CL-Contrastive Learning (or triplet, similarity loss)~ ML-Meta Learning~ PL-Prototype Learning} \\
\multicolumn{16}{l}{\textbf{\textit{Backbone(or modified):~~}}† CNN-based network defined in MADDG~\cite{shao2019multi}~ ~~‡ Dense Net~ ~ \textsuperscript{\#} ResNet-18}
\end{tabular}
}%
\vspace{-5pt}
\caption{Cross-domain evaluation on face anti-spoofing (FAS). For a fair comparison, the results of SOTA specialized methods are cited from their original papers. \textbf{Best results}, \underline{second-best}.}
\label{tab:fas}
\end{table*}

\noindent\textbf{Setting} To evaluate the generalizability of our method across diverse DfD scenarios, we follow the challenging cross-dataset setup. Specifically, we train \textit{one detector} on the FaceForensics++ (FF++, c23/HQ version)~\cite{rossler2019faceforensics++} dataset and test it on unseen datasets: CelebDF-v2 (CDFv2)~\cite{li2020celeb}, Deepfake Detection Challenge (DFDC)~\cite{dolhansky2020deepfake}, Deepfake Detection Challenge preview (DFDCp)~\cite{dolhansky2019dee}, and Wild Deepfake (WDF)~\cite{zi2020wilddeepfake}. For a fair comparison, we report the Area Under Curve (AUC) at both frame-level and video-level, the most widely used metric for DfD, in \cref{tab:df_compare}.

\noindent\textbf{Comparison with Base Models} As shown in \cref{tab:df_compare}, our FSFM significantly outperforms all base models on unseen deepfakes at both frame and video levels: 1) It surpasses Sup(IN) baselines, including ViT-B, Xception, and EfficientNet-B4, which are common practices for DfD. This suggests that our pretrained model provides a strong initialization for the detector. 2) MIM-based MAE and ID-based DINO show comparable performance but vary across different datasets, mainly because MIM focuses on local patterns while ID operates globally~\cite{zhu2023understanding}. FSFM outperforms both MAE and DINO, indicating the effectiveness of learning both local and global facial representations. 3) Notably, FSFM, pretrained on 3M VF2, consistently surpasses MCF, which is pretrained on 20M facial images and also incorporates MIM and ID. Although MCF achieves SOTA performance in face analysis tasks, its generalization to DfD is even worse than ViT, MAE, and DINO pretrained on natural images. 4) These comparisons demonstrate that FSFM effectively learns fundamental real face representations that are sensitive to deepfakes and generalizable for detection.

\noindent\textbf{Comparison with SOTA Specialized Methods} As shown by Avg.$\Delta$Ours in \cref{tab:df_compare}, FSFM consistently outperforms all counterparts, achieving the best overall performance. Our model surpasses both transformer-based and SSL-based competitors that are task-specific. In particular, our method outperforms SSL-based NACO, a most recent work~\cite{zhang2024learning} that also learns consistent representations of real face videos, and it further combines ViT-B with CNN. Remarkably, our model performs best across unseen datasets except in video-level testing on CDFV2, where some models trained on synthetic data show slightly better results. These methods (AUNet, FPG, and LAA-Net with SBIs) simulate pseudo-fake features, especially blending artifacts similar to face-swapping in CDFV2, but struggle with forgeries lacking such clues. Moreover, our method yields notable gains on more challenging DFDC and WDF datasets, which contain diverse unknown real-world manipulations. Overall, our FSFM, a vanilla ViT-B base model, achieves SOTA performance without any specialized network modules or tailored data generation for deepfake detection.

\subsection{Face anti-spoofing}
\label{subsec:exp_fas}

\noindent\textbf{Setting} To evaluate the transferability of our method for FAS under significant domain shifts, we use four widely used benchmark datasets: MSU-MFSD (M)~\cite{wen2015face}, CASIA-FASD (C)~\cite{zhang2012face}, Idiap Replay-Attack (I)~\cite{chingovska2012effectiveness}, and OULU-NPU (O)~\cite{boulkenafet2017oulu}. We treat each dataset as the target domain and apply the leave-one-out (LOO) cross-domain evaluation. As prior work~\cite{huang2022adaptive} also finetunes vanilla ViT-B for this protocol, we follow its 0-shot setting by initializing weights from our FSFM and base models. We report mean HTER (Half Total Error Rate) and AUC over 5 runs in \cref{tab:fas}. 

\begin{table}
\centering
\resizebox{\linewidth}{!}{%
\begin{tabular}{p{2.6cm}p{1.8cm}<{\centering}cccccc} 
\hline
\multirow{2}{*}{Method} & \multirow{2}{*}{\begin{tabular}[c]{@{}c@{}}Pretrain\\or Init\end{tabular}} &  & \multicolumn{4}{c}{Test Subset (AUC\%↑)} & \multirow{2}{*}{\begin{tabular}[c]{@{}c@{}}Avg. w/o\\FF++\end{tabular}} \\ 
\cline{3-7}
 &  & FF++ & T2I~~ & ~I2I & FS & FE &  \\ 
\hline
ViT-B~\cite{dosovitskiy2020image} & Scratch & 92.02 & 62.19 & 69.99 & 60.87 & 67.30 & 65.09 \\
ViT-B~\cite{dosovitskiy2020image} & Sup(IN) & 99.15 & 33.38 & 35.83 & 52.20 & 55.42 & 44.21 \\
MAE~\cite{he2022masked} ViT-B & SSL(IN) & 99.25 & 33.01 & 32.88 & 47.77 & 58.70 & 43.09 \\
DINO~\cite{caron2021emerging} ViT-B & SSL(IN) & 99.30 & 33.85 & 36.02 & 60.37 & 63.18 & 48.35 \\
MCF~\cite{wang2023toward} ViT-B & SSL(LFc) & \textbf{99.39} & 39.09 & 38.67 & 34.35 & 56.02 & 42.03 \\
\rowcolor[rgb]{0.8,0.8,0.8} \textbf{FSFM ViT-B} & SSL(FF++\_o) & 99.31 & 61.74 & \textbf{71.91} & \textbf{71.31} & \textbf{78.98} & \textbf{70.99} \\ 
\hline
\end{tabular}}
\vspace{-5pt}
\caption{Cross-dataset evaluation on the DiFF benchmark~\cite{cheng2024diffusion}. All models are finetuned only on FF++\_DeepFake (c23) subset~\cite{rossler2019faceforensics++}.}

\label{tab:DiFF}
\end{table}

\noindent\textbf{Comparison with Base Models} Our FSFM achieves the best average performance, with several key observations: 1) Simple finetuning on ViT-B Sup(IN) enhances FAS generalization, as noted in~\cite{huang2022adaptive, zhou2024test}. 2) General SSL methods, including MIM-based MAE~\cite{he2022masked} and ID-based DINO~\cite{caron2021emerging}, perform worse than ViT-B Sup(IN), with MAE showing notable declines, echoing the claims in~\cite{yu2024rethinking}. 3) With large-scale facial data for SSL, MCF outperforms MAE and DINO for FAS. However, similar to DfD, it still falls short of the ViT-B Sup(IN) baseline in average metrics, once again underscoring the gap between normal face analysis and security tasks. 4) Overall, our FSFM improves the generalizability of ViT for cross-domain FAS, effectively modeling credible features of live (real) faces.

\noindent\textbf{Comparison with SOTA Specialized Methods} Our FSFM ViT-B exhibits superior performance over domain generalization (DG) FAS methods using various CNNs. Notably, this comparison adheres to the baseline~\cite{huang2022adaptive}: although supplementary data (CelebA-Spoof~\cite{zhang2020celeba}) is included for finetuning ViT-B, no auxiliary supervision or DG techniques, \eg, depth map (DM) or adversarial learning (AL), are employed—only a standard cross-entropy loss. Additionally, we follow other ViT-based FAS methods~\cite{zhou2024test, liao2023domain, wang2022learning} without including CelebA-Spoof, with comparisons provided in the supplementary material. 

\subsection{Diffusion facial forgery detection}
\label{subsec:exp_diff}

\noindent\textbf{Setting} To further investigate the adaptability of our method against emerging unknown facial forgeries, we adopt cross-distribution testing using the recently released DiFF~\cite{cheng2024diffusion} benchmark. This dataset contains high-quality face images synthesized by 13 SOTA diffusion models across four subsets: T2I (Text-to-Image), I2I (Image-to-Image), FS (Face Swapping), and FE (Face Editing). We train only on the FF++\_DeepFake (c23) subset and report AUC results on the DiFF testing subsets. This evaluation is more challenging than typical DfD (\cref{subsec:exp_deeepfake}), as both the unseen manipulations and generative models are significantly different. 

\noindent\textbf{Comparison} In \cref{tab:DiFF}, our model significantly outperforms other ViT detectors.
Pretrained base models perform even worse than the model trained from scratch. This arises from severe overfitting to the deepfake distribution, which hinders discerning diffusion-generated faces. In contrast, our model benefits from fundamental representations of real faces that are not specific to forgery types. The substantial improvement in this new task highlights the robustness of our method in out-of-distribution scenarios.

\subsection{Ablation studies}
\label{subsec:exp_ablation}

\begin{table}[!t]
\centering
\resizebox{1\linewidth}{!}{%
\begin{tabular}{p{3.5cm}<{\centering}p{0.4cm}<{\centering}p{0.4cm}<{\centering}p{0.4cm}<{\centering}p{1.3cm}<{\centering}p{1.3cm}<{\centering}p{1.2cm}<{\centering}p{1.2cm}<{\centering}} 
\hline
\multirow{2}{*}{Component} & \multirow{2}{*}{$C^{1}$} & \multirow{2}{*}{$C^{2}$} & \multirow{2}{*}{$C^{3}$} & \multicolumn{2}{c}{Deepfake Detection} & \multicolumn{2}{c}{Face Anti-spoofing~} \\ 
\cline{5-8}
 &  &  &  & F-AUC↑ & V-AUC↑ & HTER↓ & AUC↑ \\ 
\hline
\multicolumn{8}{l}{\textbf{MAE~\&Masking Strategy~(w/o ID Network)}} \\
\&Random (base MAE) &  &  &  & 74.19 & 79.51 & 19.05 & 87.42~ \\
\&Fasking-I~\cite{cai2023marlin} &  &  &  & 73.80 & 78.33 & \textbf{17.81} & 87.75 \\
\&FRP & $\checkmark$ &  &  & 75.43 & 81.21 & 17.96 & 87.61 \\
\&CRFR-R &  & $\checkmark$ &  & 75.01 & 80.70 & 18.28 & 87.34 \\
\&CRFR-P & $\checkmark$ & $\checkmark$ &  & \textbf{76.11} & \textbf{81.58} & \uline{17.85} & \textbf{88.11} \\ 
\hline
\multicolumn{8}{l}{\textbf{ID \&Target View (w/ MAE\&CRFR-P)}} \\
\&Visible & $\checkmark$ & $\checkmark$ &  & 75.54 & 81.50 & 18.22 & 87.95 \\
\&Masked & $\checkmark$ & $\checkmark$ &  & 76.35 & 81.86 & 18.41 & 87.77 \\
\textbf{\&Full (FSFM)} &{\cellcolor[rgb]{0.8,0.8,0.8}}$\checkmark$ & ${\cellcolor[rgb]{0.8,0.8,0.8}}\checkmark$ &{\cellcolor[rgb]{0.8,0.8,0.8}}$\checkmark$ & \textbf{76.39} & \textbf{82.31} & \textbf{17.44} & \textbf{88.26} \\ 
\hline \hline
Design & \multicolumn{3}{c}{Setting} & \multicolumn{1}{l}{} & \multicolumn{1}{l}{} & \multicolumn{1}{l}{} & \multicolumn{1}{l}{} \\ 
\hline
\multirow{5}{*}{\begin{tabular}[c]{@{}c@{}}Online\&Target \\ \textbf{Rep Decoder}\\ ($D_{o}^r$\&$D_{t}^r$)\\ \textbf{Blocks}\end{tabular}}  & \multicolumn{3}{c}{0 \& 0} & 75.63 & 81.48 & 18.37 & 86.77 \\
 & \multicolumn{3}{c}{2 \& 0} & 75.74 & 81.14 & 18.54 & 87.22 \\
 & \multicolumn{3}{c}{1 \& 1} & 75.06 & 80.68 & 18.86 & 87.64 \\
 & \multicolumn{3}{c}{{\cellcolor[rgb]{0.8,0.8,0.8}}2 \& 2} & \textbf{\textbf{76.39}} & \textbf{\textbf{82.31}} & \textbf{\textbf{17.44}} & \textbf{\textbf{88.26}} \\
 & \multicolumn{3}{c}{3 \& 3} & 75.08 & 80.71 & 17.93 & 87.80 \\ 
\hline
\multirow{3}{*}{\begin{tabular}[c]{@{}c@{}}Online\&Target\\ \textbf{Data} ($I_v$\&$I$)\\ \textbf{Augmentation}\end{tabular}} & \multicolumn{3}{c}{crop+flip\&none} & 75.93 & 81.54 & 18.24 & 87.04 \\
 & \multicolumn{3}{c}{none\&crop+flip} & 73.39 & 78.80 & 19.13 & 86.11 \\
 & \multicolumn{3}{c}{{\cellcolor[rgb]{0.8,0.8,0.8}} none\&none} & {\textbf{76.39}} & \textbf{82.31} & \textbf{17.44} & \textbf{88.26} \\ 
\hline
\multirow{3}{*}{\begin{tabular}[c]{@{}c@{}}\textbf{Loss}\\for ID\end{tabular}} & \multicolumn{3}{c}{InfoNCE} & 75.10 & 80.60 & 18.24 & 87.37 \\
 & \multicolumn{3}{c}{MSE (\cite{grill2020bootstrap}-like)} & 74.19 & 79.34 & 18.09 & 88.21 \\
 & \multicolumn{3}{c}{{\cellcolor[rgb]{0.8,0.8,0.8}}Asym. \cref{eq:loss_id}} & \textbf{76.39} & \textbf{82.31} & \textbf{17.44} & \textbf{88.26} \\
\hline\hline
Pretraining dara & \multicolumn{3}{c}{Size} & \multicolumn{1}{l}{} & \multicolumn{1}{l}{} & \multicolumn{1}{l}{} & \multicolumn{1}{l}{} \\ 
\hline
\multirow{3}{*}{\begin{tabular}[c]{@{}c@{}}\textbf{Data size}\\(images)\end{tabular}} & \multicolumn{3}{c}{FF++\_o (0.1M)} & 76.39 & 82.31 & 17.44 & 88.26 \\
& \multicolumn{3}{c}{YTF(0.6M)~\cite{wolf2011face}} & 79.51 & 83.86 & 16.23 & 93.13 \\ 
& \multicolumn{3}{c}{VF2 (3.3M)} & \textbf{84.00} & \textbf{87.89} & \textbf{4.81} & \textbf{98.68} \\ 
\hline
\end{tabular}}
\vspace{-7pt}
\caption{Ablations on deepfake detection (DfD) and face anti-spoofing (FAS) with averaged metrics. If not specified, the model is pretrained on FF++\_o. Default settings are shaded in gray.}
\label{tab:ablation}
\end{table}

In ablations, unless otherwise stated, we pretrained the ViT-B/16 on the FF++\_o dataset, which includes $\sim$0.11M real face images from 720 training and 140 validation videos of FF++ (c23) YouTube subset~\cite{rossler2019faceforensics++}. We present the average generalization results on DfD and FAS in \cref{tab:ablation}.

\noindent\textbf{Effect of 3C} We first evaluate facial masking strategies in the vanilla MAE. Both our preliminary FRP and CRFR-R outperform random masking, validating the significance of intra-region consistency ($C^{1}$) and inter-region coherency ($C^{2}$), respectively. Further, our CRFR-P emerges as the most effective, showing that both $C^{1}$ and $C^{2}$ are critical and complementary for robust facial representation. Building on MAE\&CRFR-P, we incorporate the ID network with varied target views (\cref{fig:target_view}). The improved performance with \&Full confirms the benefit of local-to-global correspondence ($C^{3}$) guided by complete facial semantics. 

\noindent\textbf{Effect of Key Designs} \textbf{1) Blocks of $\mathbf{D_{o}^r}$\&$\mathbf{D_{t}^r}$} A lightweight 2-layer Rep decoder in both the online and target branches (2\&2) achieves a better balance between complexity and generalization, outperforming 1\&1 and 3\&3. Omitting the Rep decoder (0\&0) or adding it only to the online branch (2\&0) degrades performance, supporting the importance of bridging feature distribution in a disentangled space. \textbf{2) Data Augmentation} FSFM performs best without data augmentation (none\&none). Unlike many SSL methods, applying simple augmentation in the MIM network (crop+flip\&none) or target view (none\&crop+flip) reduces performance, indicating that CRFR-P masking introduces sufficient spatial regularization and that preserving target representations from original faces aids robustness. \textbf{3) Loss for ID Network} Pretraining with asymmetric negative cosine similarity (NCS) \ie \cref{eq:loss_id} proves more effective than the BYOL-like~\cite{grill2020bootstrap} MSE and the widely used InfoNCE, as the latter pushes different real faces (negative samples) apart that may not be suitable for face security tasks. 

\noindent\textbf{Effect of Data Scaling} As expected, pretraining FSFM on a larger dataset yields substantial gains, which suggests that incorporating more diverse real faces in open-ended environments can further improve generalization.

\subsection{Qualitative analysis}
\label{subsec:exp_vis}

\begin{figure}[t!]
\centering
\includegraphics[width=1\linewidth]{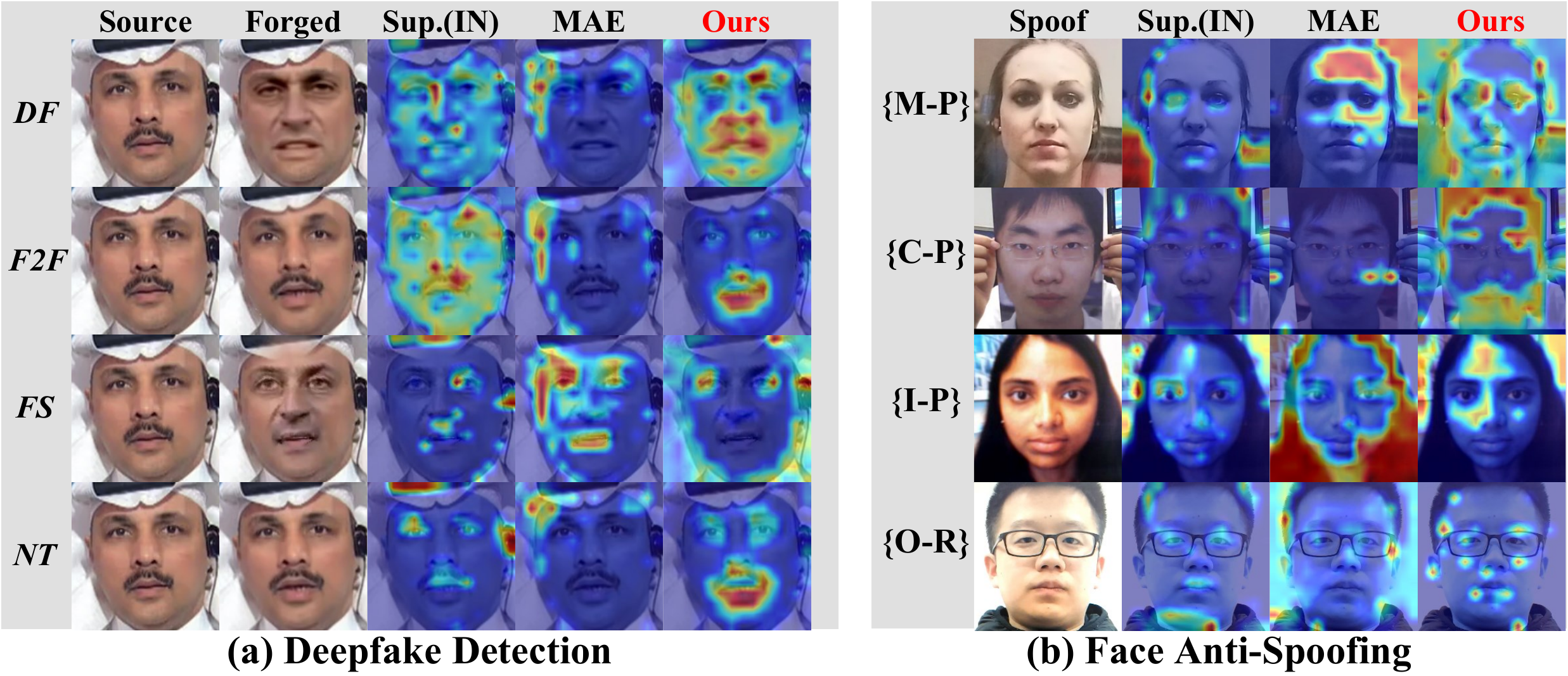}
\vspace{-17pt}
\caption{\label{fig:vis_cam} CAM Visualization. (a) DfD on various manipulations from FF++~\cite{rossler2019faceforensics++}. (b) FAS on the MCIO protocol. FSFM highlights forgery artifacts and spoofing clues. Images are from the test set.}
\end{figure}

To illustrate the superiority of the facial features learned by FSFM, we visualize GradCAM~\cite{selvaraju2017grad} activation maps for detecting forgery and spoofing, compared to Sup(IN) and MAE baselines. \textbf{1) DfD} In \cref{fig:vis_cam} (a), FSFM heatmaps more accurately identify forgery-relevant artifacts in FF++~\cite{rossler2019faceforensics++}, \eg, mouth-modified F2F and NT. \textbf{2) FAS} In \cref{fig:vis_cam} (b), FSFM predominantly captures spoof-specific clues in the MCIO cross-domain testings: high-frequency paper textures (M-Paper), photo cut edges (C-Photo), inconsistent reflections between facial regions (I-Print), and moiré patterns in video (O-Replay). These visualizations show that FSFM effectively responds to anomalies diverging from the suggested \textit{\textbf{3C}} in real/live faces, shedding light on improving the generalization performance of face security tasks.
\section{Conclusion}
In this work, we propose a self-supervised pretraining framework to learn fundamental and general representations of real faces, FSFM. We leverage the synergy between masked image modeling and instance discrimination to empower both local and global perception, promoting intra-region consistency, inter-region coherency, and local-to-global correspondence. We show that our FSFM transfers better than previous pretrained models on several face security tasks, including cross-dataset deepfake detection, cross-domain face anti-spoofing, and unseen diffusion face forgery detection. Notably, as a vanilla ViT, our model surpasses SOTA task-specific methods on generalization performance. Finally, we hope that our work can contribute to the face security and analysis community.
\section*{Acknowledgements}


This work was supported in part by the National Key R\&D Program of China under Grant 2023YFB2904000 and 2023YFB2904001, in part by the National Natural Science Foundation of China under Grant 62032021, U2436206, 62372406, and 62372402, and in part by the Zhejiang Provincial Natural Science Foundation of China under Grant LZ25F020005 and LD24F020010. The authors would like to thank all anonymous reviewers for their insightful comments on this paper. Besides, the first author wishes to thank his late grandmother, Juxian Huang, for her spiritual encouragement at the beginning of this work.
\section*{Supplementary Material}
\setcounter{section}{0}
\renewcommand{\thesection}{\Alph{section}}

\section{Overview}
This supplementary material provides additional insights, details, and results to support our FSFM framework comprehensively, structured as follows:

\noindent$\bullet$ Facial Masking Strategies in Masked Image Modeling (MIM) (\cref{sec:sup_reveal}): We delve into the impact of different facial masking strategies on naive MAE, including quantitative and qualitative analysis of attention differences.

\noindent$\bullet$ Instance Discrimination (ID) in FSFM (\cref{sec:sup_id}): We highlight the connections and distinctions between our method and prior works that also integrate ID (or Siamese encoder architecture) into MIM (or degraded inputs).

\noindent$\bullet$ Implementation Details (\cref{sec:sup_imp}): Detailed descriptions of hyperparameters, pretraining, and finetuning settings.

\noindent$\bullet$ Additional Results and Comparisons (\cref{sec:sup_exp}): Extended experiments comparing FSFM against other models like ViT-based FAS and the base vision-language pertaining (VLP) model, CLIP.

\noindent$\bullet$ Ablations and Visualizations (\cref{sec:sup_abla}): Supplementary evidence validating FSFM’s ability to learn robust and transferable facial representations.

\noindent$\bullet$ Limitations (\cref{sec:limit}) of our work.

\section{Revealing the secrets of facial masking strategies in MIM}
\label{sec:sup_reveal}

\begin{figure*}[htb!]
\centering
\includegraphics[width=.95\textwidth]{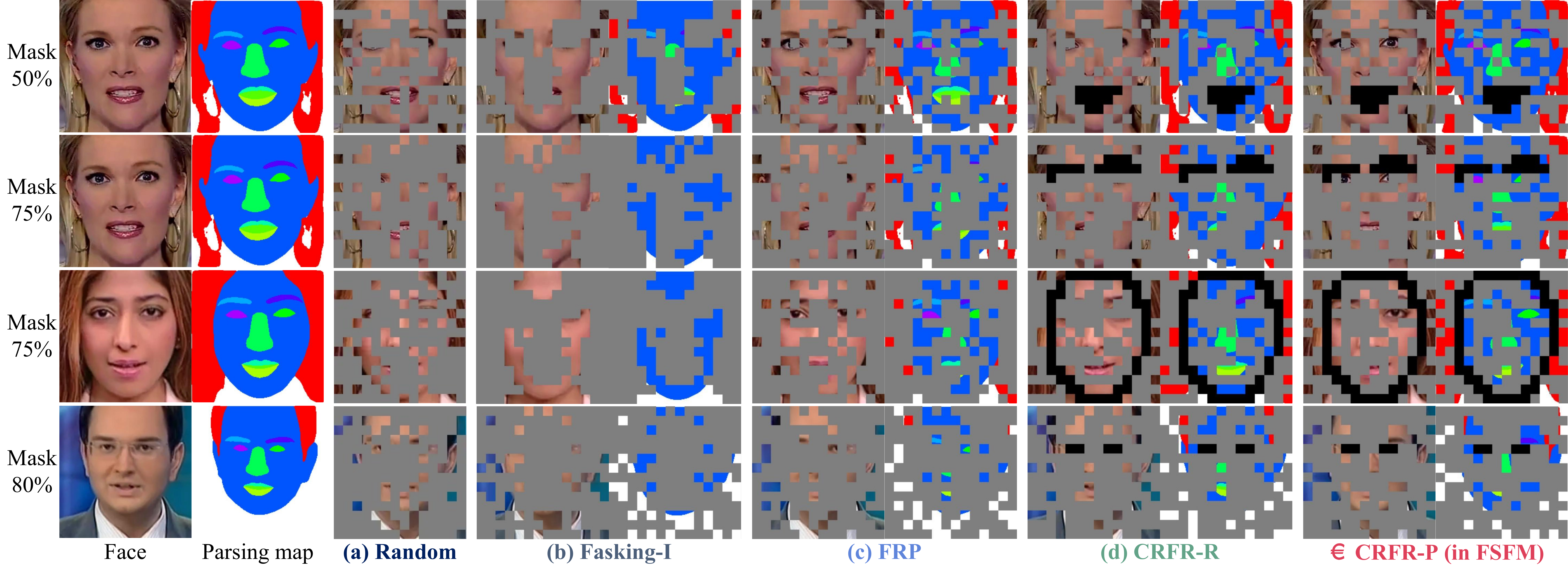}
\vspace{-5pt}
\caption{Additional visualizations of different facial masking strategies. (a) Random masking~\cite{he2022masked}. (b) Fasking-I adapted from~\cite{cai2023marlin}. (c) FRP: Facial Region Proportional masking. (d) CRFR-R: Covering a Random Facial Region followed by Random masking. (e) CRFR-P: Covering a Random Facial Region followed by Proportional masking.}
\label{fig:facial_masks}
\end{figure*}

\begin{figure*}[htb!]
\centering
\includegraphics[width=.95\linewidth]{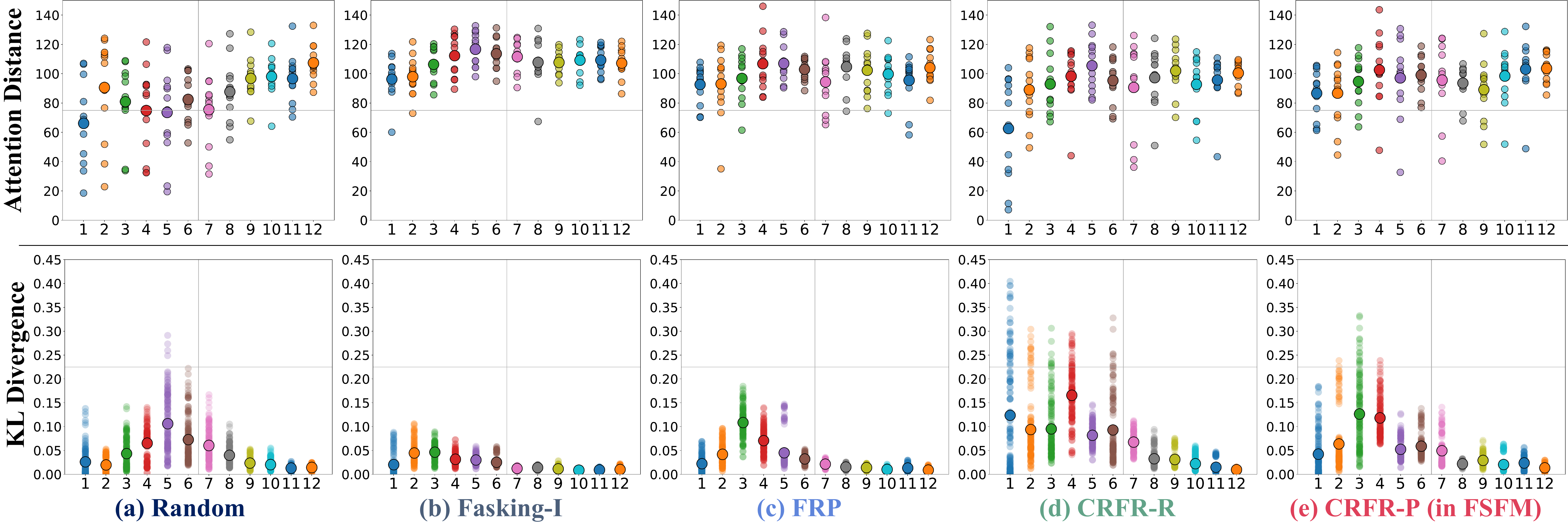}
\vspace{-5pt}
\caption{The mean attention distance (\textit{Top}) and Kullback-Leibler divergence (\textit{Bottom}) of each attention head (small dot) across all blocks (\textit{x-axis}) in the ViT-B/16 encoder pretrained by MAE~\cite{he2022masked} with (a) Random, (b) Fasking-I, (c) FRP, (d) CRFR-R, and (e) CRFR-P masking strategies, along with the average one (large dot) for each block.}
\label{fig:mask_analysis} 
\end{figure*}

\begin{figure*}[htb!]
\centering
\includegraphics[width=.95\linewidth]{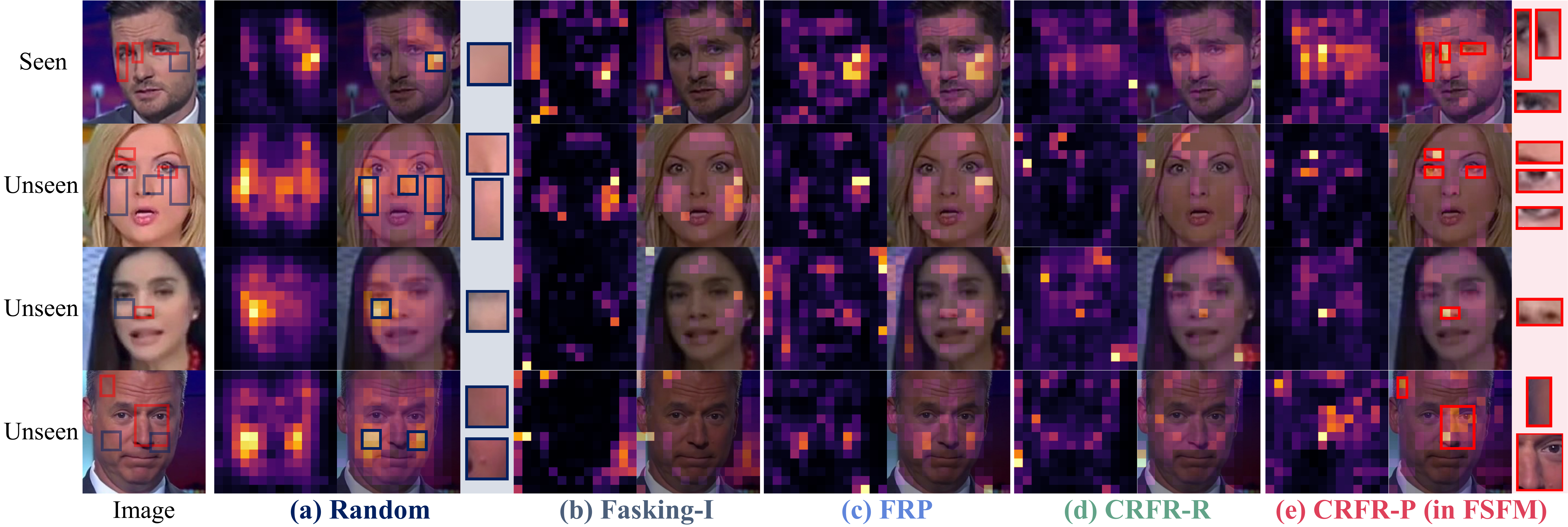}
\vspace{-5pt}
\caption{Visualization of the self-attention map averaged across all heads from the last block of the ViT-B/16 encoder pretrained by MAE~\cite{he2022masked} with (a) Random, (b) Fasking-I, (c) FRP, (d) CRFR-R, and (e) CRFR-P masking strategies. The rectangles in \textcolor{RMcolor}{(a) Random} and \textcolor{CRFR-Pcolor}{(e) CRFR-P} highlight the regions of interest (ROI) for comparison. All faces, except for the first row, are from the test set and were unseen during pretraining.}
\label{fig:att_map_mask_stra}
\end{figure*}

\begin{figure*}[htb!]
\centering
\includegraphics[width=.95\linewidth]{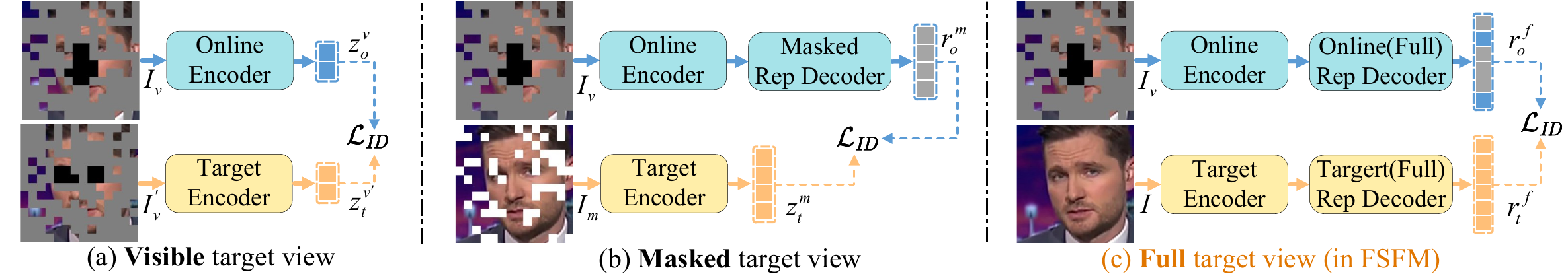}
\vspace{-5pt}
\caption{Comparison of typical target views and designs adapted for FSFM, derived from self-supervised pretraining methods that integrate both MIM and ID (or Siamese encoder architecture). (a) Visible patches from a different mask~\cite{li2023mage, wang2023toward, bai2023masked, zhao2024asymmetric}. (b) Masked patches from the same mask~\cite{chen2024context}. (c) Full patches without masking~\cite{zhou2021ibot, yi2022masked, assran2022masked, huang2023contrastive, tao2023siamese}.}
\label{fig:target_view_all}
\end{figure*}

In the main paper, we explore various facial masking strategies for masked image modeling (MIM) in FSFM, with additional visualizations provided in \cref{fig:facial_masks}, and validate the effectiveness of CRFR-P masking through ablation studies on downstream face security tasks. However, a critical question remains: how do different facial masking strategies affect the MIM-pretrained model or its learned representations of real faces?

To address this, we quantitatively and qualitatively analyze the properties of attention maps. Given that most MIM-pretrained models, including ours, are built on the Vision Transformer (ViT) architecture~\cite{dosovitskiy2020image}, where the main component, the attention mechanism~\cite{vaswani2017attention}, is naturally interpretable~\cite{xie2023revealing}. Here, we adopt the naive MAE (the MIM network in FSFM) and follow its settings~\cite{he2022masked} with ViT-B/16 as the encoder and a 75\% masking ratio. We conduct self-supervised pretraining on real face images from FF++\_o~\cite{rossler2019faceforensics++} (the default dataset for our ablations). We alter only the masking strategy, encompassing simple random masking, Fasking-I, FRP, CRFR-R, and CRFR-P, and examine the following aspects of attention heads in the pretrained models: 1) mean attention distance to measure the flow of local and global facial information; 2) Kullback-Leibler (KL) divergence to investigate the diversity of attention; 3) visualized attention maps to uncover key regions of focus.

\subsection{Local or global?}
To explore whether the pretrained model attends to faces locally or globally, we calculate the mean attention distance~\cite{dosovitskiy2020image} in each attention head across all blocks/layers, as shown in \cref{fig:mask_analysis} (\textit{Top}). The model (MAE ViT-B/16 encoder) pretrained with simple random masking tends to focus on local information in the lower blocks and shifts toward global attention in the deeper blocks, similar to the supervised model~\cite{dosovitskiy2020image}. Fasking-I primarily aggregates global information as the visible patches predominantly consist of broad backgrounds and skin. FRP also causes large mean attention distances, but these are slightly smaller than those of Fasking-I, mainly because visible patches in FRP are more evenly distributed across all facial regions. CRFR-R fully masks a facial region before applying random masking, which encourages attention to different regions, consequently resulting in more global attention in the middle (3\textsuperscript{rd} to 8\textsuperscript{th}) blocks compared to the simple random masking counterparts. Compared with CRFR-R, CRFR-P masks the remaining regions proportionally instead of randomly, making the 1\textsuperscript{st} block more global \wrt the more unmasked regions. Compared with FRP, CRFR-P fully masks a region before applying proportional masking, which exposes more patches within other regions at the same masking ratio, thus leading to more local attention than FRP.

Overall, the model pretrained with CRFR-P exhibits well-distributed attention distances across all blocks, indicating a synergistic effect of FRP and CRFR-R, enabling appropriate attention to both local and global information.

\subsection{Similar or different?}
To assess whether the pretrained model focuses on similar or different tokens, we compute the Kullback-Leibler (KL) divergence between attention maps of each head across all blocks, following~\cite{xie2023revealing}, as shown in \cref{fig:mask_analysis} (\textit{Bottom}). As the visible patches are mostly background and skin, the model pretrained with Fasking-I aggregates similar tokens, leading to low KL divergence across all attention heads. Interestingly, we find that proportional masking reduces diversity among attention heads, likely due to its homogeneous presentation of visible tokens, \ie, derived from each facial region. In contrast, covering a random facial region increases attention diversity, as evidenced by higher KL divergence in CRFR-R versus the simple random masking counterparts and CRFR-P versus the FRP counterparts. This suggests that the model is compelled to look at different regions after fully masking a facial region.

Overall, the model pretrained with FRP lacks diversity across attention heads, while CRFR-R shows excessive diversity. Similar to the phenomenon observed in mean attention distance, CRFR-P strikes a balance in KL divergence across different heads, which also seems to act as a co-effect of FRP and CRFR-R counterparts, implicating appropriate attention to different key tokens.

\subsection{Key regions?}
To uncover which regions of real faces are critical to the pretrained model, we visualize the mean attention map of the last block and overlay it onto the input face in \cref{fig:att_map_mask_stra}, as the pretext decoder or downstream head follows the last block. We can observe significant attention differences in salient regions across different masking strategies. At first glance, the attention regions from simple random masking appear to cover the entire face. However, it predominantly highlights the skin, which can be easily recovered from visible neighboring patches, while ignoring more challenging key regions. This suggests that the model solves face reconstruction through shortcuts instead of learning meaningful features. Similarly, the attention in Fasking-I is distributed across skin and background regions, as expected. While FRP and CRFR-R activate more attention areas, they still struggle to focus on key facial regions. In contrast, CRFR-P highlights attention across key regions like the eyes, eyebrows, and nose, indicating that the pretrained model tackles the challenge head-on: focusing on meaningful region-level features beyond just low-level pixels of real faces.

In summary, the CRFR-P masking strategy effectively directs attention to critical facial regions with appropriate range and diversity for both intra-region consistency and inter-region coherency, enabling the pretrained facial model to avoid trivial solutions (shortcuts) and capture the intrinsic properties of real faces. Furthermore, we hope this section provides new insights into fundamental face representation.

\section{Connection and analysis of ID in FSFM}
\label{sec:sup_id}
We illustrate the relation and distinction between our method and previous works that integrate ID (or Siamese encoder architecture) into MIM (or degraded input). While these hybrid approaches have demonstrated effectiveness in natural vision and face analysis, our empirical studies reveal that face security tasks necessitate more precise and reliable semantic alignment from the ID network. In response, we distinguish key designs such as target view \& network structure, data augmentation, and loss function, which support local-to-global correspondences in FSFM.

\subsection{Target view and network design}
From the perspective of the input view, the online/student branch typically processes visible patches from the masked image, while the target/teacher branch varies across methods. Thus, we incorporate different target views and design paradigms into FSFM, as shown in \cref{fig:target_view_all}. (a) Visible patches from a different mask~\cite{li2023mage, wang2023toward, bai2023masked, zhao2024asymmetric}: the online and target encoders produce latent features $z_o^v$ and $z_t^v$ for subsequent contrast learning. (b) Masked patches from the same mask~\cite{chen2024context}: to align the representation $r_o^m$ of masked patches with the encoded target $z_t^m$, a masked representation (rep) decoder predicts $r_o^m$ from the visible tokens $z_o^v$ output by the online encoder. This decoder computes cross-attention between masked tokens (as Q) and full tokens (as K and V), following the latent regressor in CAE~\cite{chen2024context} and resembling the prompting decoder in~\cite{hou2022milan}. (c) Full patches without masking~\cite{zhou2021ibot, yi2022masked, assran2022masked, huang2023contrastive, tao2023siamese}: some methods~\cite{zhou2021ibot, yi2022masked, assran2022masked} are decoder-free designs that match visible online features $z_o^v$ with full target features $z_o^f$ to compute $\mathcal{L}_{ID}(z_o^v, z_t^f)$. Unlike CMAE~\cite{huang2023contrastive}, which introduces a feature/rep decoder after the online encoder, \ie, $\mathcal{L}_\mathit{ID}(r_o^f, z_t^f)$, we add an additional target rep decoder to compute $\mathcal{L}_\mathit{ID}(r_o^f,r_t^f)$ in a disentangled representation space. \ie, Siamese rep decoders. This design further reduces the gap in distribution between low-level pixel features and high-level semantic representations.

Based on our ablations (in the main paper) of downstream face security tasks, FSFM performs better when using full patches as the target view alongside Siamese rep decoders. By predicting the entire face representation from visible parts, the ID network aligns global and local views of the same face. In light of this, FSFM structures the encoded space with semantically complete facial representations through ``local-to-global'' correspondences~\cite{caron2021emerging}, which endows the encoder with strong facial discriminability.

\subsection{Data augmentation}
Most ID methods~\cite{chen2020simple, he2020momentum, chen2021empirical, grill2020bootstrap, chen2021exploring, caron2021emerging} rely on strong data augmentations, including spatial and color transforms, to prevent trivial solutions. For MIM, applying strong augmentations such as color enhancements is suboptimal~\cite{he2022masked}, as masking corruption itself effectively regularizes the pretext task. This is further evident in methods~\cite{yi2022masked, li2023mage, huang2023contrastive} that combine MIM and ID, where only simple augmentations---random size cropping or flipping---are applied to the masked input of the online branch, while strong or simple augmentations are used for the full (unmasked) target view.

In contrast, our FSFM behaves well without any data augmentation in either the online or target branches. This may stem from the semantic integrity preserved in unaugmented inputs, which benefits the learning of global face identity~\cite{wang2023toward}, especially in face security tasks where forgery and spoofing cues may be implicit anywhere. Additionally, the proposed CRFR-P masking strategy inherently introduces spatial variance tailored to facial structures, rendering simple augmentations (crop and flip) unnecessary. Consequently, FSFM processes only a single view per face image.

\subsection{Loss function for ID}
We consider two main types of loss functions for ID: contrastive~\cite{chen2020simple, chen2021empirical} and non-contrastive~\cite{grill2020bootstrap, chen2021exploring}. Contrastive loss pulls positive views from the same sample together and pushes negative views from different samples apart. We use the widely adopted InfoNCE~\cite{oord2018representation} as the contrastive loss. Non-contrastive loss maximizes the similarity of positive representations only. We use mean squared error (MSE) in BYOL~\cite{grill2020bootstrap} and negative cosine similarity (NCS) in SimSiam~\cite{chen2021exploring} as non-contrastive loss, respectively, but in an asymmetric formulation, as detailed in the main paper.

In FSFM, we observe that NCS outperforms the contrastive loss InfoNCE, even though methods combining MIM and ID~\cite{li2023mage, yi2022masked, huang2023contrastive, wang2023toward} favor the latter. We speculate this is because, in large-scale pretraining on real faces, the inter-image contrast introduced by negative sample pairs—---pushing one real face away from others---does not help our model learn representations beneficial for face security tasks. Thus, we adopt asymmetric NCS by default to learn intra-face correspondences by matching the online anchor view with the target view of the same sample.

\begin{table}[t!]
\centering
\resizebox{\linewidth}{!}{%
\begin{tabular}{lccccc} 
\hline
\multirow{2}{*}{Base Model} & \multirow{2}{*}{\begin{tabular}[c]{@{}c@{}}Pretrain\\or Init\end{tabular}} & \multicolumn{2}{c}{Intra-FF++ (c23)~} & \multicolumn{2}{c}{Intra-FF++ (c40)} \\ 
\cline{3-6}
 &  & F-AUC & V-AUC & F-AUC & V-AUC \\ 
\hline
ViT-B~\cite{dosovitskiy2020image} & Scratch & 70.06 & 72.98 & 70.36 & 74.33 \\
ViT-B~\cite{dosovitskiy2020image} & Sup(IN) & 97.30 & 98.80 & 88.65 & 93.22 \\
MAEE~\cite{he2022masked} ViT-B & SSL(LN) & \textbf{98.64} & \textbf{99.61} & \uline{91.35} & \uline{94.46} \\
DINO~\cite{caron2021emerging} ViT-B & SSL(LN) & \uline{98.35} & \uline{99.39} & 89.93 & 94.31 \\
MCF~\cite{wang2023toward} ViT-B & SSL(LFc) & 97.84 & 99.19 & 89.81 & 94.20 \\
\rowcolor[rgb]{0.8,0.8,0.8} FSFM ViT-B (Ours) & SSL(VF2) & 97.74 & 99.03 & \textbf{92.08} & \textbf{95.41} \\
\hline
\end{tabular}}
\vspace{-3pt}
\caption{Intra-dataset evaluation of deepfake detection (DfD) on FF++~\cite{rossler2019faceforensics++}. All base models are finetuned and tested on the c23 and c40 versions, respectively.  \textbf{Best results}, \underline{second-best}.}
\label{tab:intra_FF}
\end{table}

\begin{table*}[htb]
\centering
\resizebox{\linewidth}{!}{
\begin{tabular}{p{5.2cm}p{1.8cm}<{\centering}p{0.4cm}<{\centering}p{0.4cm}<{\centering}p{0.4cm}<{\centering}p{0.4cm}<{\centering}p{0.4cm}<{\centering}p{1.3cm}<{\centering}p{1.3cm}<{\centering}p{1.3cm}<{\centering}p{1.3cm}<{\centering}p{1.3cm}<{\centering}p{1.3cm}<{\centering}p{1.3cm}<{\centering}p{1.3cm}<{\centering}p{1.3cm}<{\centering}p{1.3cm}<{\centering}}
\hline
\multirow{2}{*}{Method} & \multirow{2}{*}{\begin{tabular}[c]{@{}c@{}}Pretrain\\or Init\end{tabular}} & \multicolumn{5}{c}{DG FAS Technique} & \multicolumn{2}{c}{OCI→M} & \multicolumn{2}{c}{OMI→C} & \multicolumn{2}{c}{~OCM→I} & \multicolumn{2}{c}{ICM→O} & \multicolumn{2}{c}{Avg.} \\ 
\cline{3-17}
 &  & DM & AL & CL & ML & PL & HTER↓ & AUC↑ & HTER↓ & AUC↑ & HTER↓ & AUC↑ & HTER↓ & AUC↑ & HTER↓ & AUC↑ \\ 
\hline
\multicolumn{17}{l}{\textbf{Base model}} \\
ViT-B~\cite{dosovitskiy2020image}\textbf{*} & Scratch &  &  &  &  &  & 15.25 & 90.07 & 36.98 & 63.94 & 10.75 & 96.09 & 37.50 & 66.03 & 25.12 & 79.03~ \\
ViT-B~\cite{dosovitskiy2020image}\textbf{*} & Sup(IN) &  &  &  &  &  & \uline{5.00} & 97.29 & 11.51 & 95.61 & 10.60 & 94.03 & 17.35 & 88.35 & 11.12 & 93.82~ \\
MAE~\cite{he2022masked} ViT-B\textbf{*} & SSL(LN) &  &  &  &  &  & 8.17 & 97.37 & 27.91 & 77.90 & 18.02 & 91.54 & 25.36 & 80.76 & 19.86 & 86.89~ \\
DINO~\cite{caron2021emerging} ViT-B\textbf{*} & SSL(LN) &  &  &  &  &  & 7.92 & 97.09 & 21.28 & 87.79 & 21.35 & 84.48 & 17.44 & 89.75 & 17.00 & 89.78~ \\
MCF~\cite{wang2023toward} ViT-B\textbf{*} & SSL(LFc) &  &  &  &  &  & 6.33 & \uline{98.29} & 21.40 & 86.72 & 13.13 & 95.15 & 16.76 & 89.59 & 14.41 & 92.44~ \\
\rowcolor[rgb]{0.8,0.8,0.8} \textbf{FSFM ViT-B (Ours)}\textbf{*} & SSL(VF2) &  &  &  &  &  & 6.58 & 97.43 & \textbf{4.30} & \textbf{99.10} & 14.63 & 90.96 & \uline{10.02} & \textbf{96.23} & \uline{8.88} & \uline{95.93} \\ 
\hline
\hline
\multicolumn{17}{l}{\textbf{ViT-based specialized method (Venue)}} \\
ViTranZFAS~\cite{george2021effectiveness} (IJCB'21)\textbf{*} & Init(IN) &  &  &  &  &  & 10.95 & 95.05 & 14.33 & 92.10 & 16.64 & 85.07 & 15.67 & 89.59 & 14.40 & 90.45~ \\
TransFAS~\cite{wang2022face} (TBIOM'22)\textbf{*} & Init(IN) & \checkmark &  &  &  &  & 7.08 & 96.69 & 9.81 & 96.13 & 10.12 & 95.53 & 15.52 & 91.10 & 10.63 & 94.86~ \\
TTN-S~\cite{wang2022learning} (TIFS'22)\textbf{*} & Init(IN) & \checkmark &  &  &  &  & 9.58 & 95.79 & 9.81 & 95.07 & 14.15 & 94.06 & 12.64 & 94.20 & 11.55 & 94.78~ \\
DiVT-V~\cite{liao2023domain} (WACV'23)\textbf{*} & Init(IN) &  &  & \checkmark & \checkmark &  & 10.00 & 96.64 & 14.67 & 93.08 & \textbf{5.71} & \uline{97.73} & 18.06 & 90.21 & 12.11 & 94.42~ \\
TTDG-V~\cite{zhou2024test} (CVPR'24)\textbf{*} & Init(IN) &  &  & \checkmark &  &  & \textbf{4.16} & \textbf{98.48} & \uline{7.59} & \uline{98.18} & \uline{9.62} & \textbf{98.18} & \textbf{10.00} & \uline{96.15} & \textbf{7.84} & \textbf{97.75} \\
\hline
\multicolumn{16}{l}{\textit{\textbf{Abbreviation:~~}}Sup-Supervised~ SSL-Self-Supervied~ Init-weight initialization~ IN-ImageNet~ DG-Domain Generalization} \\
\multicolumn{16}{l}{\textit{\textbf{DG FAS Technique:~~}}DM-Depth Maps~ AL-Adversarial Learning~ CL-Contrastive Learning (or triplet, similarity loss)~ ML-Meta Learning~ PL-Prototype Learning}
\end{tabular}
}%
\vspace{-5pt}
\caption{Cross-domain evaluation on face anti-spoofing (FAS) using visual-only ViT-based models\textbf{*} without including the supplementary data in~\cite{huang2022adaptive}. For a fair comparison, the results of specialized methods are cited from the original papers. \textbf{Best results}, \underline{second-best}.}
\label{tab:fas-vit}
\end{table*}

\begin{table*}[htb]
\centering
\resizebox{\linewidth}{!}{%
\begin{tabular}{lccccccc|lccccccc} \hline
\multirow{2}{*}{Method} & \multirow{2}{*}{\begin{tabular}[c]{@{}c@{}}Pretrain \\or Init\end{tabular}} & \multirow{2}{*}{\begin{tabular}[c]{@{}c@{}}Train\\Set\end{tabular}} & \multicolumn{4}{c}{Test Set \textbf{Video-level} AUC(\%)$\uparrow$} & \multirow{2}{*}{\begin{tabular}[c]{@{}c@{}}Avg.\\$\Delta$Ours\end{tabular}} & \multirow{2}{*}{Method} & \multirow{2}{*}{\begin{tabular}[c]{@{}c@{}}Pretrain \\or Init\end{tabular}} & \multirow{2}{*}{\begin{tabular}[c]{@{}c@{}}Train\\Set\end{tabular}} & \multicolumn{4}{c}{Test Set \textbf{Frame-level} AUC(\%)$\uparrow$} & \multirow{2}{*}{\begin{tabular}[c]{@{}c@{}}Avg.\\$\Delta$Ours\end{tabular}} \\ 
\cline{4-7}\cline{12-15}
\multicolumn{1}{c}{} &  &  & CDFV2 & DFDC & DFDCP & WDF &  & \multicolumn{1}{c}{} &  &  & CDFV2 & DFDC & DFDCP & WDF &  \\ 
\hline
\multicolumn{8}{l|}{\textbf{Base model}} & \multicolumn{8}{l}{\textbf{Base model}} \\
Xception~\cite{chollet2017xception} & Sup(IN) & FF++ & 76.39 & 70.62 & 72.24 & 76.11 & 14.0↑ & Xception~\cite{chollet2017xception} & Sup(IN) & FF++ & 69.52 & 68.20 & 68.94 & 68.83 & 15.1↑ \\
EfficientNet-B4~\cite{tan2019efficientnet} & Sup(IN) & FF++ & 79.81 & 71.85 & 66.95 & 76.42 & 14.1↑ & EfficientNet-B4~\cite{tan2019efficientnet} & Sup(IN) & FF++ & 73.37 & 69.47 & 64.37 & 71.95 & 14.2↑ \\
ViT-B~\cite{dosovitskiy2020image} & Scratch & FF++ & 64.08 & 66.73 & 72.62 & 60.36 & 21.9↑ & ViT-B~\cite{dosovitskiy2020image} & Scratch & FF++ & 61.14 & 64.27 & 69.00 & 60.68 & 20.2↑ \\
ViT-B~\cite{dosovitskiy2020image} & Sup(IN) & FF++ & \uline{86.24} & 74.48 & 82.11 & 81.20 & 6.9↑ & ViT-B~\cite{dosovitskiy2020image} & Sup(IN) & FF++ & \uline{77.43} & 71.09 & 74.07 & \uline{75.86} & 9.4↑ \\
MAE~\cite{he2022masked} ViT-B & SSL(IN) & FF++ & 79.51 & 75.93 & \uline{87.10} & 80.96 & 7.0↑ & MAE~\cite{he2022masked} ViT-B & SSL(IN) & FF++ & 72.64 & 72.18 & \uline{79.81} & 73.94 & 9.4↑ \\
DINO~\cite{caron2021emerging} ViT-B & SSL(IN) & FF++ & 80.47 & \uline{76.90} & 84.64 & \uline{82.06} & 6.9↑ & DINO~\cite{caron2021emerging} ViT-B & SSL(IN) & FF++ & 73.88 & \uline{72.78} & 77.31 & 75.08 & 9.2↑ \\
MCF~\cite{wang2023toward} ViT-B & SSL(LFc) & FF++ & 80.25 & 73.61 & 82.55 & 79.79~ & 8.8↑ & MCF~\cite{wang2023toward} ViT-B & SSL(LFc) & FF++ & 73.16 & 69.63 & 75.78 & 74.10 & 10.8↑  \\
CLIP~\cite{radford2021learning} ViT-B & VLP(WIT) & FF++ & 78.95 & 73.83 & 82.38 & 78.60 & 9.5↑ & CLIP~\cite{radford2021learning} ViT-B & VLP(WIT) & FF++ & 73.02 & 70.66 & 77.46 & 72.04 & 10.7↑  \\
\rowcolor[rgb]{0.8,0.8,0.8} \textbf{FSFM ViT-B~\textbf{(Ours)}} & SSL(VF2) & FF++ & \textbf{91.44} & \textbf{83.47} & \textbf{89.71} & \textbf{86.96} & - & \textbf{FSFM ViT-B (Ours)} & SSL(VF2) & FF++ & \textbf{85.05} & \textbf{80.20} & \textbf{85.50} & \textbf{85.26} & - \\ 
\hline
\multicolumn{16}{l}{\textit{\textbf{Abbreviation:~~}}Sup-Supervised~ SSL-Self-Supervied~ VLP-Vision Language pretraining~ Init-weight initialization} \\
\multicolumn{16}{l}{\textit{\textbf{Dataset:~~}}IN/1M natural images~\cite{deng2009imagenet}~ LFc/20M facial images~\cite{wang2023toward}~ WIT/400M (natural image, text) pairs~\cite{radford2021learning}~ VF2/3M facial images~\cite{cao2018vggface2}}
\end{tabular}
}%
\vspace{-5pt}
\caption{Cross-dataset evaluation on deepfake detection (DfD), \textit{adding CLIP~\cite{radford2021learning} ViT-B/16 image encoder as a base model}. Left: video-level AUC, Right: frame-level AUC. All base models are finetuned on FF++ (c23) and tested on unseen datasets. Avg.$\Delta$Ours denotes the average AUC difference between our FSFM and other methods. \textbf{Best results}, \underline{second-best}.}
\label{tab:df_compare_clip}
\end{table*}

\section{More implementation details}
\label{sec:sup_imp}
\subsection{Pretraining settings}
\label{sec:dps}

We set the mask ratio $r$ to 0.75, similar to the baseline~\cite{he2022masked}, as our ablation shows that this high ratio is also favorable for our FSFM. We do not use any data augmentation (not even crop and flip used in~\cite{he2022masked}) and only normalize the input faces during pretraining. We empirically set the loss weights $\lambda_\mathit{fr}$ and $\lambda_\mathit{cl}$ to 0.007 and 0.1, respectively. The projection and prediction heads are 2-layer MLPs following BYOL~\cite{grill2020bootstrap}, with batch normalization (BN) replaced by layer normalization (LN) for our ViT-based architecture. The EMA momentum coefficient for updating the target branch starts from 0.996 and increases with a cosine scheduler, following BYOL~\cite{grill2020bootstrap}. We pretrain our model from scratch for 400 epochs on 4 NVIDIA RTX A6000 GPUs. Other settings follow the defaults in MAE~\cite{he2022masked}: we use the AdamW~\cite{loshchilov2017decoupled} optimizer with momentum $\beta_1 = 0.9$, $\beta_2 = 0.95$; we apply the linear lr scaling rule~\cite{goyal2017accurate} with a base learning rate of 1.5e-4; we adopt a cosine decay~\cite{loshchilov2016sgdr} learning rate schedule with a warmup epoch of 40; we maintain the effective batch size as 4096 = 256 (batch size per GPU) $\times$ 4 (GPUs) $\times$ 4 (accumulated gradient iterations).

\subsection{Finetuning settings in downstream tasks}
\label{sec:ps}

For finetuning the ViTs from FSFM and other pretrained models, we adopt identical settings except for weight initialization, detailed below.

\noindent\textbf{Deepfake Detection} We use the c23 (HQ) version of FF++~\cite{rossler2019faceforensics++} with official train/val splits for finetuning. We sample 128 frames per real video (the original YouTube subset) and 32 frames per forgery video (including Deepfakes, Face2Face, FaceSwap, and NeuralTextures subsets). We follow the official test split in other unseen datasets for testing, including CDFV2~\cite{li2020celeb}, DFDC~\cite{dolhansky2020deepfake}, DFDCp~\cite{dolhansky2019dee}, and WDF~\cite{zi2020wilddeepfake}. We sample 32 frames per testing video. We use DLIB~\cite{king2009dlib} to extract faces (without alignment and parsing) and resize them to 224$\times$224. As WDF already provides 224$\times$224 facial images, we directly use its test set without processing. We add only one linear layer as the binary classifier after averaging all non-[CLS] token features. We set the batch size to 64, the base learning rate to 2.5e-4, and the finetuning epochs to 10 (50 for ViT-B Scratch). Other settings adhere to the MAE ImageNet finetuning recipe~\cite{he2022masked}.

\begin{table*}[htb]
\centering
\resizebox{.7\linewidth}{!}{%
\begin{tabular}{lccccccccccc} \hline
\multirow{2}{*}{Method} & \multirow{2}{*}{\begin{tabular}[c]{@{}c@{}}Pretrain\\or Init\end{tabular}} & \multicolumn{2}{c}{OCI→M} & \multicolumn{2}{c}{OMI→C} & \multicolumn{2}{c}{OCM→I} & \multicolumn{2}{c}{ICM→O} & \multicolumn{2}{c}{Avg.} \\ \cline{3-12}
 &  & HTER↓ & AUC↑ & HTER↓ & AUC↑ & HTER↓ & AUC↑ & HTER↓ & AUC↑ & HTER↓ & AUC↑ \\ \hline
\multicolumn{12}{l}{\textbf{Base model}} \\
ViT-B~\cite{dosovitskiy2020image} & Scratch & 15.37 & 90.73 & 35.37 & 68.23 & 14.75 & 94.18 & 31.65 & 71.55 & 24.28 & 81.17~ \\
ViT-B~\cite{dosovitskiy2020image, huang2022adaptive} & Sup(IN) & \textbf{3.52} & 98.74 & \textbf{2.42} & \textbf{99.52} & 8.45 & 96.91 & 11.86 & 94.62 & 6.56 & 97.44 \\
MAE~\cite{he2022masked} ViT-B & SSL(LN) & 10.32 & 94.87 & 15.91 & 89.96 & 15.54 & 91.13 & 16.51 & 90.29 & 14.57 & 91.56 \\
DINO~\cite{caron2021emerging} ViT-B & SSL(LN) & 6.73 & 97.15 & 13.44 & 93.90 & 14.27 & 93.56 & 15.55 & 90.99 & 12.50 & 93.90 \\
MCF~\cite{wang2023toward} ViT-B & SSL(LFc) & 4.00 & \uline{98.84} & 8.46 & 96.90 & 8.02 & 97.39 & 10.70 & 95.64 & 7.80 & 97.19 \\
CLIP~\cite{radford2021learning} ViT-B & VLP(WIT) & 4.29 & 98.76 & 5.00 & 98.89 & \uline{7.14} & \uline{97.92} & \textbf{6.09} & \textbf{98.12} & \uline{5.63} & \uline{98.42} \\
\rowcolor[rgb]{0.8,0.8,0.8} \textbf{FSFM ViT-B (Ours)} & SSL(VF2) & \uline{3.78} & \textbf{99.15} & \uline{3.16} & \uline{99.41} & \textbf{4.63} & \textbf{99.03} & \uline{7.68} & \uline{97.11} & \textbf{4.81} & \textbf{98.68} \\ \hline
\multicolumn{12}{l}{\textit{\textbf{Abbreviation:~~}}Sup-Supervised~ SSL-Self-Supervied~ VLP-Vision Language pretraining~ Init-weight initialization} \\
\multicolumn{12}{l}{\textit{\textbf{Dataset:~~}}IN-ImageNet1K~\cite{deng2009imagenet}~ LFc-LAION FACE cropped~\cite{wang2023toward}~ WIT-WebImageText~\cite{radford2021learning}~ VF2-VGGFace2~\cite{cao2018vggface2}}
\end{tabular}
}%
\vspace{-5pt}
\caption{Cross-domain evaluation on face anti-spoofing (FAS), \textit{adding CLIP~\cite{radford2021learning} ViT-B/16 image encoder as a base model}. \textbf{Best}, \underline{second}.}
\label{tab:fas_clip}
\end{table*}

\begin{table}[htb]
\centering
\resizebox{\linewidth}{!}{%
\begin{tabular}{p{2.6cm}p{1.7cm}<{\centering}cccccc} 
\hline
\multirow{2}{*}{Method} & \multirow{2}{*}{\begin{tabular}[c]{@{}c@{}}Pretrain\\or Init\end{tabular}} &  & \multicolumn{4}{c}{Test Subset (AUC\%↑)} & \multirow{2}{*}{\begin{tabular}[c]{@{}c@{}}Avg. w/o\\FF++\end{tabular}} \\ 
\cline{3-7}
 &  & FF++ & T2I~~ & ~I2I & FS & FE &  \\ 
\hline
ViT-B~\cite{dosovitskiy2020image} & Scratch & 92.02 & \uline{62.19} & \uline{69.99} & 60.87 & \uline{67.30} & \uline{65.09} \\
ViT-B~\cite{dosovitskiy2020image} & Sup(IN) & 99.15 & 33.38 & 35.83 & 52.20 & 55.42 & 44.21 \\
MAE~\cite{he2022masked} ViT-B & SSL(IN) & 99.25 & 33.01 & 32.88 & 47.77 & 58.70 & 43.09 \\
DINO~\cite{caron2021emerging} ViT-B & SSL(IN) & 99.30 & 33.85 & 36.02 & 60.37 & 63.18 & 48.35 \\
MCF~\cite{wang2023toward} ViT-B & SSL(LFc) & \textbf{99.39} & 39.09 & 38.67 & 34.35 & 56.02 & 42.03 \\
CLIP~\cite{radford2021learning} ViT-B & VLP(WIT) & \uline{99.33} & \textbf{69.63} & 66.25 & \uline{65.23} & 57.07 & 64.54 \\
\rowcolor[rgb]{0.8,0.8,0.8} \textbf{FSFM ViT-B} & SSL(FF++\_o) & 99.31 & 61.74 & \textbf{71.91} & \textbf{71.31} & \textbf{78.98} & \textbf{70.99} \\
 \hline
\end{tabular}}
\vspace{-3pt}
\caption{Cross-dataset evaluation on DiFF benchmark~\cite{cheng2024diffusion}. \textit{adding CLIP~\cite{radford2021learning} ViT-B/16 image encoder as a base model}. All base models are finetuned only on the FF++\_DeepFake (c23)~\cite{rossler2019faceforensics++}. \textbf{Best results}, \underline{second-best}.}
\label{tab:DiFF_clip}
\end{table}

\noindent\textbf{Face Anti-Spoofing} In the main paper, we adopt the 0-shot MCIO setting (Protocol 1) in~\cite{huang2022adaptive} and include CelebA-Spoof~\cite{zhang2020celeba} as supplementary data for FAS finetuning. We set the batch size to 12 for each training domain. We append the MLP head after averaging all non-[CLS] token features instead of using [CLS] ones~\cite{huang2022adaptive}, to align with other face security tasks. Additionally, for a fair comparison with other visual-only ViT-based methods, we additionally follow \cite{zhou2024test} and report the best performance without including the supplementary data, as presented in \cref{sec:fas-vit}.

\noindent\textbf{Diffusion Face Forgery Detection} For the training set, we sample 32 frames from each real video (the original YouTube subset) and each forgery video (the Deepfakes subset) from FF++ (c23)~\cite{rossler2019faceforensics++}. For validation and testing sets, we follow the splits provided by the DiFF benchmark~\cite{10.1145/3664647.3680797}. We use DLIB~\cite{king2009dlib} to extract faces (without alignment and parsing) and resize them to 224$\times$224. We add one linear layer as the binary classifier after averaging all non-[CLS] token features. We set the batch size to 256, the base learning rate to 5e-4, and the finetuning epochs to 50. Other settings adhere to the MAE ImageNet finetuning recipe~\cite{he2022masked}.

\section{Additional experimental results}
\label{sec:sup_exp}

\subsection{Comparison of intra-dataset DfD on FF++}
While our primary objective focuses on cross-domain generalization for real-world applicability, we provide an intra-dataset evaluation of deepfake detection (DfD) on FaceForensics++ (FF++)~\cite{rossler2019faceforensics++}, as presented in \cref{tab:intra_FF}. The metrics on the FF++\_c23 show that FSFM maintains comparable intra-set performance while significantly improving cross-dataset generalization. Moreover, when evaluated on the more challenging high-compression (c40) version, FSFM outperforms all baseline vision models, further demonstrating its robustness. 

\subsection{Comparison with ViT-based FAS}
\label{sec:fas-vit}

In a fair comparison with visual-only ViT-based face anti-spoofing (FAS) methods, our method also significantly outperforms all base models, as shown in \cref{tab:fas-vit}. FSFM surpasses most counterparts and ranks second in average metrics. TTDG-V~\cite{zhou2024test}, which introduces test-time domain generalization and explicit optimization goals for FAS, performs better than ours in two out of four target domains (OCI→M and OCM→I). While optimizing for a specific downstream task is beyond the scope of this study, incorporating special auxiliary supervision or domain generalization (DG) techniques into our pretrained model may further improve its generalization ability for face presentation attack detection.

\subsection{Comparison with CLIP}

Another line of representation learning, vision-language pretraining (VLP), particularly contrastive language-image pretraining (CLIP)~\cite{radford2021learning}, has shown remarkable zero-shot and generalization capabilities across diverse downstream tasks. Recent studies have successfully tailored CLIP to specific face security tasks, including deepfake detection~\cite{khan2024clipping, tan2024c2p, smeu2024declip}, face anti-spoofing~\cite{srivatsan2023flip, liu2024cfpl, hu2024fine, guo2024style, 10.1007/978-3-031-72897-6_10}, and diffusion forgery detection~\cite{zhang2024mfclip, cozzolino2024raising, lin2024robust}. These text-aided methods differ fundamentally from our FSFM, which is vision-only, self-supervised, and task-agnostic. Moreover, VLP demands extensive (image, text) data pairs along with significant computing resources for the additional text encoder. Despite these, we include CLIP as a base vision-language model (VLM) for comparison.

Specifically, we borrow the CLIP image encoder, also a ViT-B but pretrained on the WIT dataset with 400M image\& text pairs, and finetune it on downstream face security tasks under the same settings as other base models. We supplement the corresponding results on deepfake detection, face anti-spoofing, and diffusion face forgery detection in \cref{tab:df_compare_clip}, \cref{tab:fas_clip}, and \cref{tab:DiFF_clip}, respectively. We can observe that CLIP ViT-B transfers better than other base vision models on FAS and DiFF tasks, benefiting from the extensive data scale of multi-modal supervision. However, directly applying CLIP ViT-B to DfD exhibits inferior generalization. In contrast, our proposed FSFM consistently outperforms CLIP ViT-B across downstream face security tasks.

\section{More ablations and visualizations}
\label{sec:sup_abla}

\subsection{Ablation studies}

\begin{table}[htb]
\centering
\resizebox{.9\linewidth}{!}{%
\begin{tabular}{cccccc} \hline
\multirow{2}{*}{Component} & \multirow{2}{*}{Setting} & \multicolumn{2}{c}{Deepfake Detection} & \multicolumn{2}{c}{Face Anti-spoofing~} \\ \cline{3-6}
 &  & F-AUC↑ & V-AUC↑ & HTER↓ & AUC↑ \\ \hline
\multirow{5}{*}{\begin{tabular}[c]{@{}c@{}}Masking\\ratio $r$\end{tabular}} & 0.35 & 73.92 & 79.56 & 15.84 & 90.02 \\
 & 0.50 & 74.31 & 79.48 & 21.57 & 84.09 \\
 & 0.65 & 74.92 & 80.13 & 17.76 & 87.37 \\
 & {\cellcolor[rgb]{0.8,0.8,0.8}}0.75 & \textbf{\textbf{76.39}} & \textbf{\textbf{82.31}} & \textbf{\textbf{17.44}} & \textbf{\textbf{88.26}} \\
 & 0.85 & 75.40 & 80.83 & 19.19 & 86.13 \\ \hline \hline
Pretraining model & Size &  &  &  &  \\ \hline
\multirow{3}{*}{\begin{tabular}[c]{@{}c@{}}\textbf{Model size}\\(parameters)\end{tabular}} & ViT-S/16(22M) & 74.80 & 80.20 & 19.32 & 89.13 \\
 & {\cellcolor[rgb]{0.8,0.8,0.8}}ViT-B/16(86M) & 76.39 & 82.31 & 17.44 & 88.26 \\
 & ViT-L/16(303M) & \textbf{77.43} & \textbf{83.15} & \textbf{16.23} & \textbf{93.13} \\ \hline
\end{tabular}}
\vspace{-3pt}
\caption{Ablations on deepfake detection (DfD) and face anti-spoofing (FAS) with average metrics. The model is pretrained on FF++\_o~\cite{rossler2019faceforensics++}. Default settings are shaded in gray.}
\label{tab:ablation_supp}
\end{table}

\begin{figure}[htb]
\centering
\includegraphics[width=.9\linewidth]{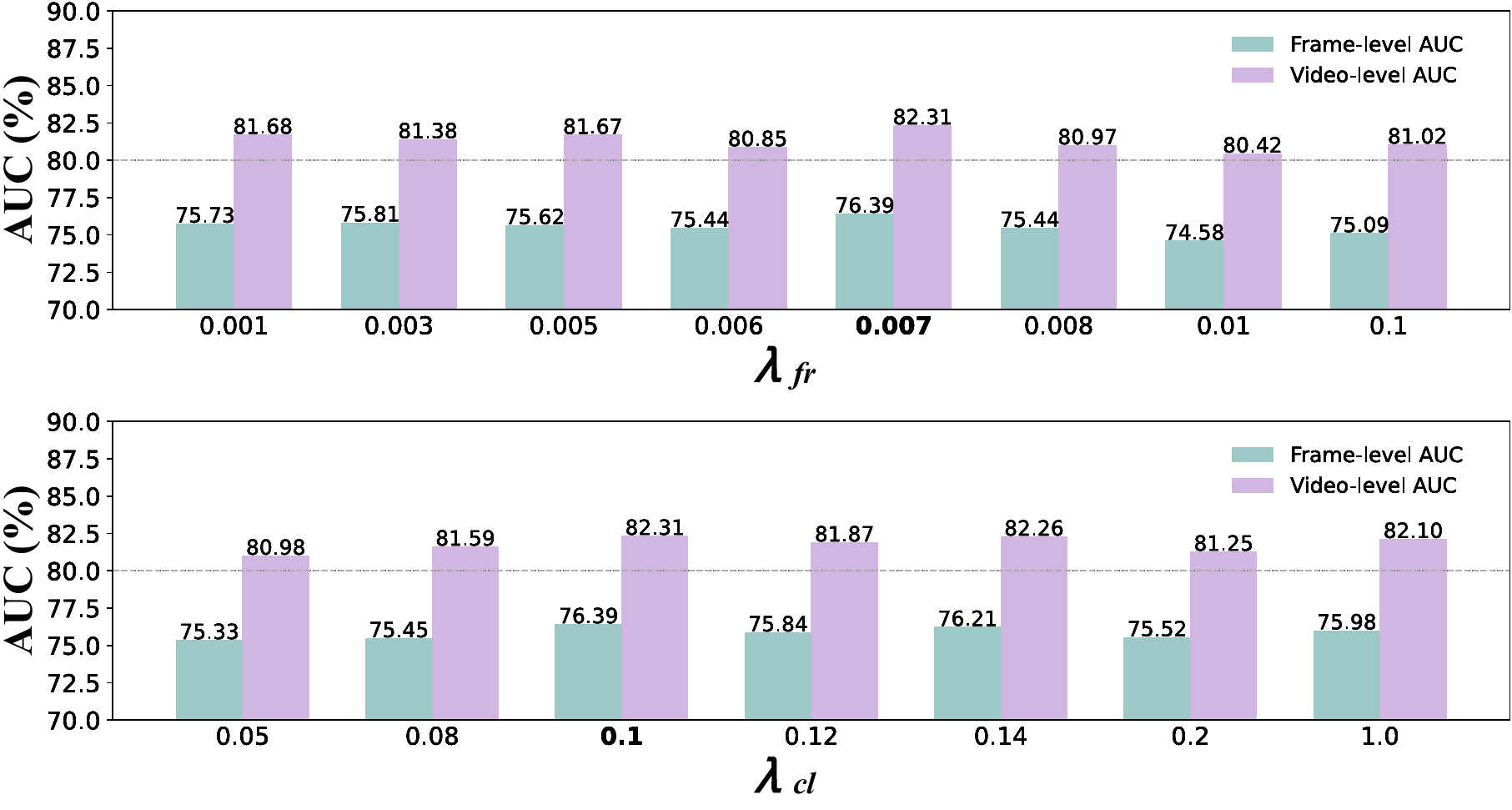}
\vspace{-5pt}
\caption{\label{fig:loss_weight} Ablations of loss weights on deepfake detection (DfD) with average metrics. The model is pretrained on FF++\_o~\cite{rossler2019faceforensics++}. Default: $\lambda_\mathit{fr}=0.007$, $\lambda_\mathit{cl}=0.1$.}
\end{figure}

\begin{figure}[htb]
\centering
\includegraphics[width=.85\linewidth]{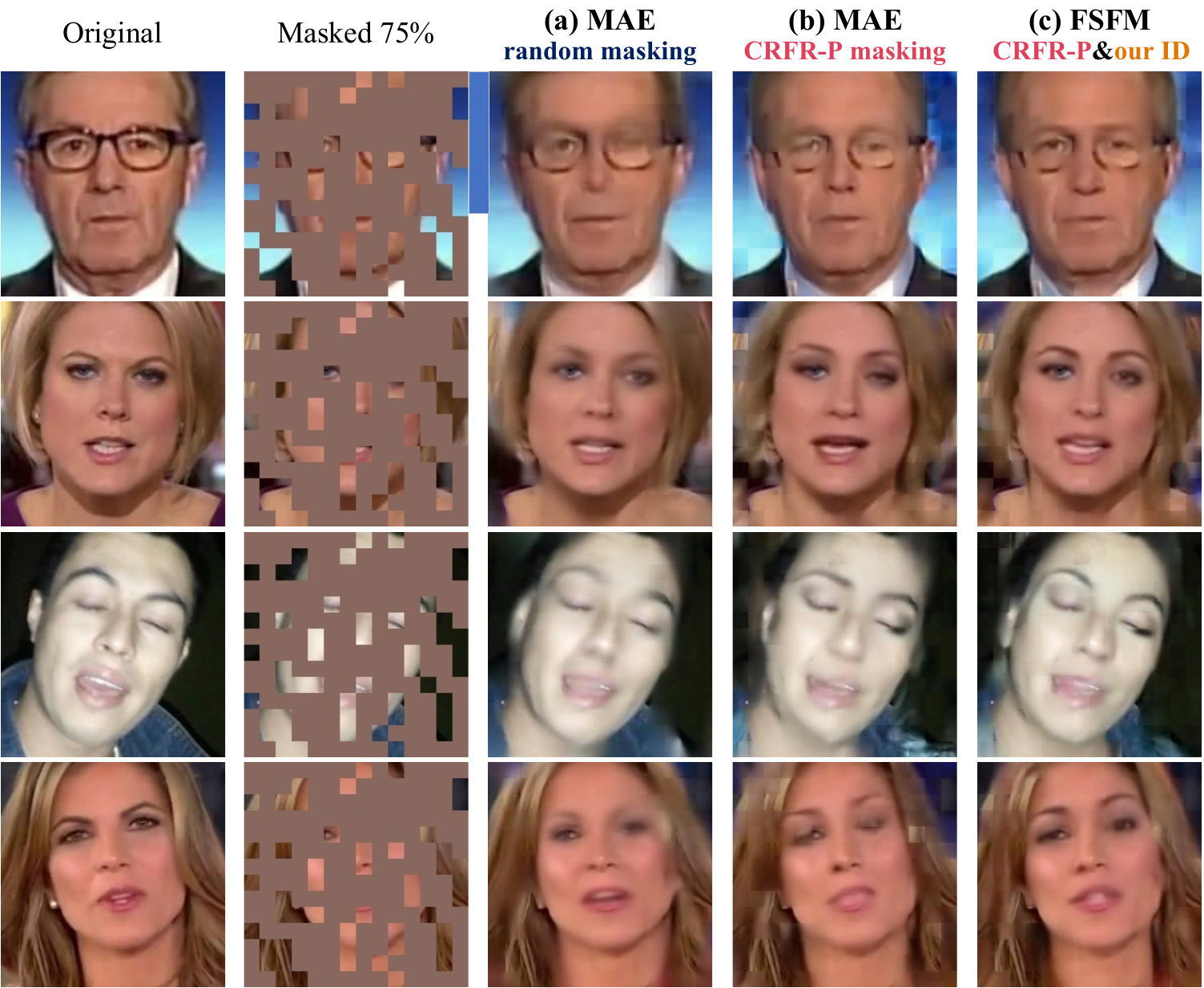}
\vspace{-5pt}
\caption{\textbf{Reconstruction Visualization} of real face images with a masking ratio of 75\%, using MIM models pretrained from: (a) a naive MAE with simple random masking~\cite{he2022masked}, (b) a naive MAE with our CRFR-P masking, and (c) our FSFM. All models were pretrained on the train and validation sets of FF++\_o~\cite{rossler2019faceforensics++} without adversarial learning, for 400 epochs. Images are from the test set.}
\label{fig:rec}
\end{figure}

This subsection presents additional ablations. Unless otherwise stated, the default settings follow the main paper.

\noindent\textbf{Effect of Masking Ratio $\mathit{\mathbf{r}}$} We also examine the impact of different masking ratios for CRFR-P masking on our pretraining framework. As shown in \cref{tab:ablation_supp}, FSFM achieves the best overall performance with a 0.75 masking ratio. Adopting lower masking ratios leads to trivial reconstruction and alignment tasks due to more available information. Conversely, using a higher masking ratio makes pretext tasks too challenging to learn sufficient facial representations for downstream face security tasks. Accordingly, we select a 75\% masking ratio as the default setting.

\noindent\textbf{Effect of Model Scaling} \cref{tab:ablation_supp} also shows that FSFM benefits from larger model sizes when pretrained on FF++\_o. The transfer performance on downstream face security tasks improves as the model scales up. Due to limited computing resources, we were unable to pretrain larger models on more face images, \eg, pertaining ViT-L/16(303M) on the full VGGFace2 dataset, but we aspire to explore this in future work and update model zoos accordingly.

\noindent\textbf{Effect of Loss Weight} To explore the impact of the reconstruction and distillation losses, we vary various loss weights, \ie $\lambda_\mathit{fr}$ and $\lambda_\mathit{cl}$. Results in \cref{fig:loss_weight} show that the configuration ($\lambda_\mathit{fr}=0.007$, $\lambda_\mathit{cl}=0.1$) performs better on challenging cross-dataset DfD.

\subsection{Visualizations}

\noindent\textbf{Reconstruction} To demonstrate the superiority of the facial representations pretrained with FSFM, we further follow MAE~\cite{he2022masked} to visualize reconstruction results, as shown in \cref{fig:rec}. We can see that FSFM demonstrates better reconstruction quality concerning intra-region consistency (preserving fine-grained textures within facial regions), inter-region coherency (maintaining spatial relationships across regions), and local-to-global correspondence (aligning local appearance with global facial looking).

\begin{figure}[t]
\centering
\includegraphics[width=1\linewidth]{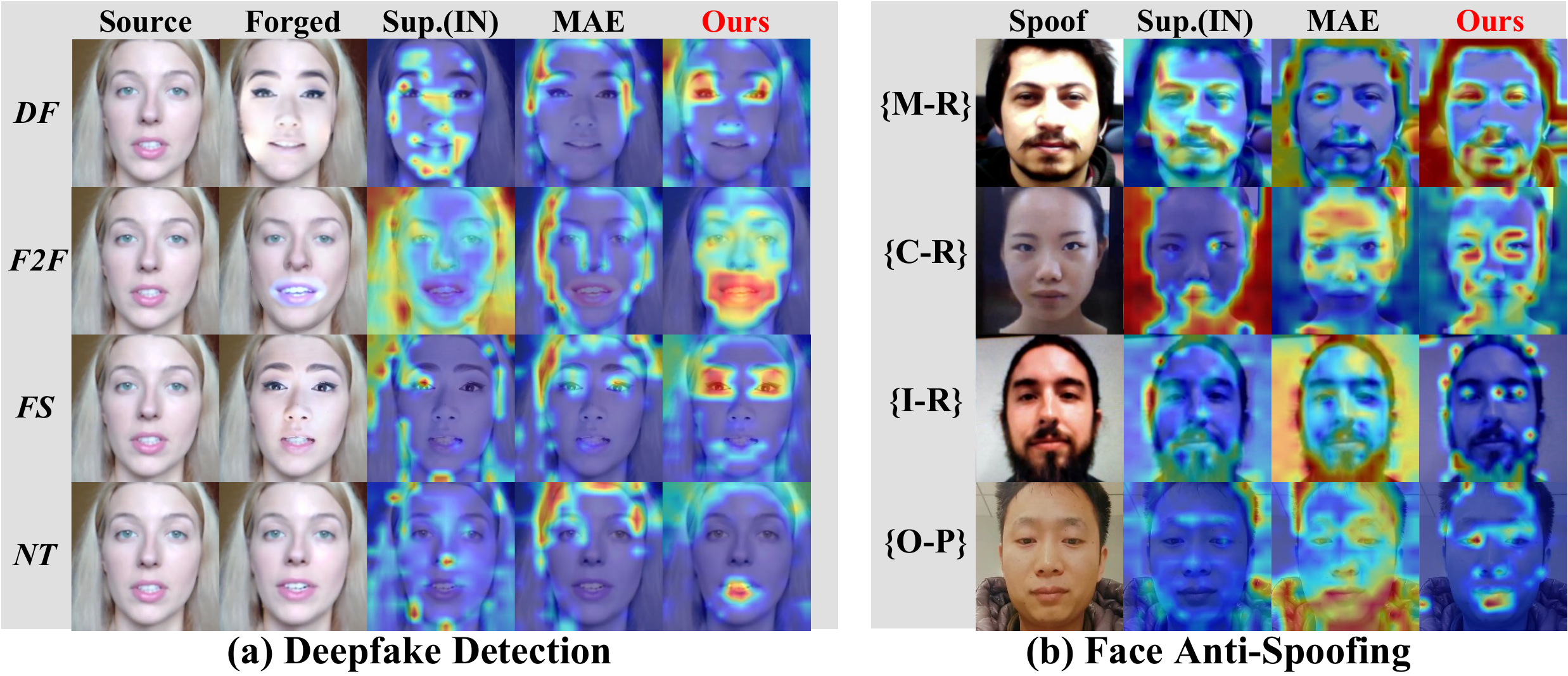}
\vspace{-15pt}
\caption{\label{fig:vis_cam_2} \textbf{CAM Visualization.} (a) DfD on various manipulations from FF++~\cite{rossler2019faceforensics++}. (b) FAS on the MCIO protocol. FSFM highlights forgery artifacts and spoof clues. Images are from the test set.}
\end{figure}

\noindent\textbf{CAM} We provide additional CAM visualizations in \cref{fig:vis_cam_2}, which are consistent with the observations in the main paper, further substantiating the effectiveness of our method.

\section{Limitations} 
\label{sec:limit}
Despite the promising results demonstrated by FSFM across various face security tasks, our work has several limitations that warrant further exploration: \textbf{Pretraining Dataset Bias} FSFM is pretrained on large-scale facial images, and its performance can be affected by the quantity, diversity, and quality of the pretraining data. Pretraining on specific datasets like VGGFace2~\cite{cao2018vggface2} may inherit their biases (\eg, race, ethnicity, and age), potentially reducing fairness. \textbf{Absence of Multi-modal Learning} Since our work focuses on general visual face security, the current framework processes only image or frame data for downstream forgery\&spoofing image\&video detection, ignoring potential complementary signals (\eg, audio inconsistencies in deepfakes or physiological cues in spoofing), which could further enhance capabilities.

{   
    \small
    \bibliographystyle{ieeenat_fullname}
    \bibliography{main}
}


\end{document}